\documentclass{article}

\usepackage{iclr2027_conference,times}

\usepackage[utf8]{inputenc}
\usepackage[T1]{fontenc}
\usepackage{hyperref}
\usepackage{url}
\usepackage{tikz}
\usepackage{booktabs}
\usepackage{amsfonts}
\usepackage{amsmath}
\usepackage{amssymb}
\usepackage{nicefrac}
\usepackage{microtype}
\usepackage{graphicx}
\usepackage{wrapfig}
\usepackage{caption}
\usepackage{multirow}
\usepackage{array}
\usepackage[table,dvipsnames]{xcolor}
\usepackage{enumitem}
\usepackage{placeins}
\usepackage{subcaption}
\usepackage{siunitx}

\title{Embedding Physical Reasoning into Diffusion-Based Shadow Generation Under the Sun and Sky}

\author{
Shilin Hu\textsuperscript{1} \quad
Jingyi Xu\textsuperscript{1} \quad
Akshat Dave\textsuperscript{1} \quad
Dimitris Samaras\textsuperscript{1}\thanks{Equal advising.} \quad
Hieu Le\textsuperscript{2}\footnotemark[1]
\\[0.5em]
\textsuperscript{1}\begin{tabular}[t]{@{}l@{}}
Department of Computer Science\\
Stony Brook University
\end{tabular}
\qquad
\textsuperscript{2}\begin{tabular}[t]{@{}l@{}}
Department of Computer Science\\
University of North Carolina at Charlotte
\end{tabular}
\\[0.5em]
{\small
\texttt{\{shilhu,jingyixu,dave,samaras\}@cs.stonybrook.edu}
\quad
\texttt{hle40@charlotte.edu}
}
}

\iclrfinalcopy % Show authors and disable review line numbers.

\begin{document}
\maketitle

% The style sets a conference header inside \maketitle; clear it afterwards.
\fancyhead{}
\renewcommand{\headrulewidth}{0pt}

\begin{abstract}

Generating realistic cast shadows for inserted foreground objects requires reasoning about scene geometry and illumination. However, most learning-based approaches treat shadow generation as an image translation problem and capture these physical relationships only implicitly. This often results in misaligned or implausible shadows. Motivated by the physics of shadow formation, we introduce explicit geometric guidance for outdoor shadow generation. Given a composite image and a foreground object mask, we recover approximate scene geometry and estimate a dominant light direction to derive a coarse shadow estimate via geometric reasoning. While coarse, this estimate provides a spatial anchor for shadow placement. Because illumination cannot always be uniquely inferred from a single image, we predict confidence scores for both lighting and shadow cues and use them to regulate their influence during generation. These cues (shadow mask, light direction, and their confidence scores) condition a diffusion-based generator that refines the estimate into a realistic shadow. Experiments on DESOBAV2 show substantially improved shadow-region fidelity and localization, with an overall 23\% lower shadow-region RMSE and 30\% lower shadow-mask BER than the prior state-of-the-art method.

\end{abstract}
\section{Introduction}
\label{sec:intro}

Shadow formation is a geometric process: given a 3D object and a light direction, light rays from the source are occluded by the object, casting a shadow whose shape and placement are physically determined on receiver surfaces. Yet modern shadow generation methods~\citep{hu_iccv2019mask, zhang2019shadowgan, Liu2020ARShadowGANSG, hong2022shadow, liu2024shadow, zhao2025shadow} primarily operate in 2D pixel space, leaving the model to implicitly discover this physical relationship from data alone. Thus, the result is often inconsistent: shadows that drift, misalign, or hallucinate, especially with complex object geometry or ambiguous lighting.

\begin{figure}[!t]
\centering
\includegraphics[width=0.75\linewidth]{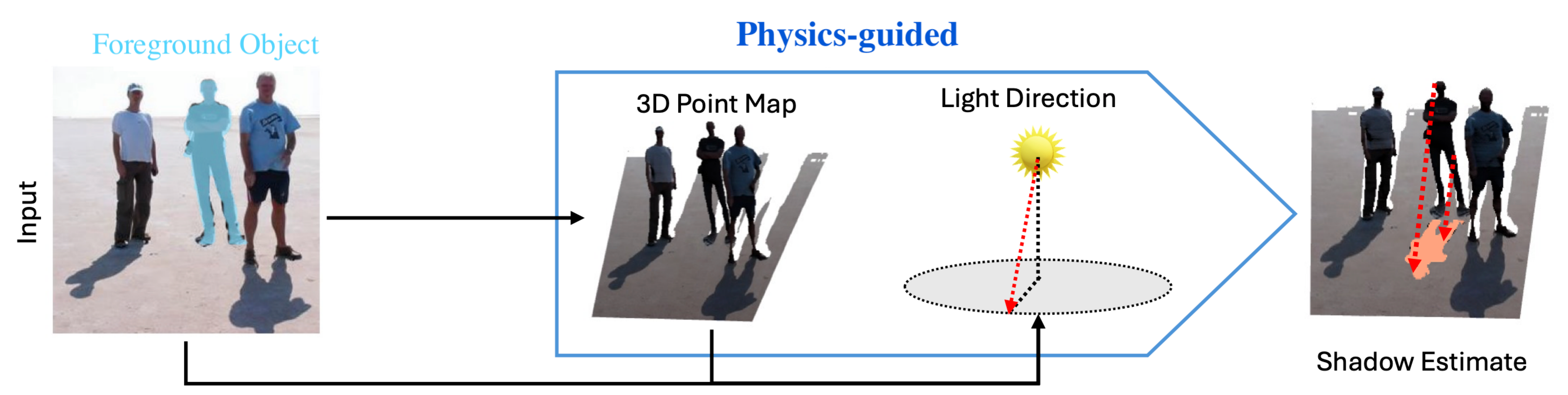}
\caption{\textbf{Shadow estimates from approximate geometry and light direction.} Given a monocular RGB image and a foreground object mask, we recover an approximate 3D point map from MoGe-2~\citep{wang2025moge2} and estimate a single dominant light direction, then infer a shadow estimate using object points. This shadow estimate serves as a condition for our shadow generator.}
\label{fig: teasersmall}
\end{figure}

Our approach instead is motivated by the physics of shadow formation: if approximate scene geometry and a light direction are known, the location and extent of a shadow can be inferred directly through geometric reasoning. Given a composite image and a mask of the foreground object whose cast shadow is absent, as illustrated in Figure~\ref{fig: teasersmall}, we recover approximate scene geometry in the form of a dense 3D point map using MoGe-2~\citep{wang2025moge2} and estimate a dominant light direction, which together allow us to compute a physics-guided shadow estimate via geometric ray reasoning~\citep{sato2003illumination, Panagopoulospami13}. Although coarse, this estimate provides an explicit spatial anchor for generation, encoding the object--receiver--illumination relationship rather than requiring the generator to learn it entirely from data.

However, lighting cannot always be uniquely inferred from an image~\citep{Lalonde09, Panagopoulospami13}. In many scenes, the visual evidence required to determine the illumination is weak or absent, for example, when no reference object–shadow pairs are visible, making the mapping from appearance to light direction inherently ill-posed. To account for this uncertainty, our framework estimates confidence scores for the predicted light direction and shadow mask and uses them to regulate how strongly each cue influences generation: reliable predictions guide the synthesis, while uncertain ones are down-weighted, allowing the model to rely more on learned image priors.

Finally, we incorporate the predicted shadow mask, light direction, and their confidence scores into a diffusion-based shadow generator~\citep{ho2020denoising,zhao2025shadow}. The shadow mask provides dense spatial conditioning for shadow placement, while the light direction supplies scene-level cues for global shadow orientation. Confidence scores modulate these signals, and the diffusion model uses the resulting physics-guided conditioning to synthesize a visually coherent shadow, recovering appearance, softness, and scene interaction.

We evaluate our method on DESOBAV2~\citep{liu2024shadow}. Experiments show that our method improves both image quality and shadow mask accuracy, producing more visually faithful and spatially consistent shadows. Compared with the prior state-of-the-art~\citep{zhao2025shadow}, our method achieves 23\% lower shadow-region RMSE and 30\% lower shadow-mask BER.  

Our contributions are summarized as follows:
\begin{itemize}[leftmargin=*]
\item \textbf{Physics-guided shadow generation.}
We introduce a framework that derives an initial shadow estimate from recovered monocular scene geometry and a predicted dominant light direction, providing a physics-guided spatial anchor for shadow placement.

\item \textbf{Quality-aware conditioning for uncertain physical cues.}
We model the reliability of predicted lighting and shadow cues with confidence scores, allowing unreliable physical guidance to be down-weighted when illumination or shadow estimates are ambiguous.

\item \textbf{Diffusion-based refinement with geometry and illumination guidance.}
We incorporate the predicted shadow mask and light direction into a generator through quality-aware spatial and scene-level conditioning, enabling realistic shadow synthesis with improved shadow localization and geometric consistency.

\item \textbf{State-of-the-art performance on DESOBAV2.}
Experiments demonstrate improved shadow-region fidelity and localization, achieving 23\% lower shadow-region RMSE and 30\% lower shadow-mask BER than the prior state-of-the-art method.
\end{itemize}

\section{Related Work}

\noindent \textbf{Image composition.}
Image composition integrates a foreground object into a background scene to produce a visually coherent composite image~\citep{niu2021making}.
Earlier work addressed individual aspects of image composition, including image blending~\citep{perez2003poisson,wu2019gpgan}, image harmonization~\citep{cong2020dovenet}, and geometric correction~\citep{lin2018stgan}.
Recent work has advanced object placement~\citep{zhou2025bootplace} and generative image composition~\citep{song2023objectstitch,song2024imprint,chen2024anydoor}.
Although such general-purpose methods may synthesize shadows as part of the composite, they typically either do not treat cast-shadow generation as a dedicated task or infer shadow placement only implicitly from image data. Our work focuses specifically on generating the missing cast shadow for a foreground object using explicit geometry and illumination cues.

\noindent \textbf{Shadow generation.}
While shadow detection~\citep{zhu18b,m_Le-etal-ECCV18,Hu2019RevisitingSD,wang2024swinshadow} and shadow removal~\citep{Cun2020TowardsGS,le2021physics,guo2023shadowdiffusion,hu2025shadow} operate on observed image content by identifying or correcting shadow regions, shadow generation instead synthesizes a cast shadow that is absent from the input.
This requires modeling both shadow appearance and placement while maintaining consistency with the foreground object, scene geometry, and illumination.
Existing shadow generation methods can be broadly categorized into 3D-based rendering approaches and 2D-based generative methods.

\noindent \textbf{3D-based methods} use explicit scene representations, including object geometry, lighting, and camera parameters, to simulate shadows through explicit rendering.
Recovering these factors from a single image is an ill-posed inverse-rendering problem because geometry, illumination, and material appearance are tightly entangled.
Early techniques~\citep{chan2003rendering,schwarz2007bitmask} used depth maps and known light sources to produce soft shadows in screen space, while \citet{sheng2022controllable} used a manually provided pixel height map and user-specified lighting for controllable shadow generation.
These methods provide explicit physical control but require detailed scene representations or manual input, limiting their use in image composition.

\noindent \textbf{2D-based generative methods} synthesize shadows directly in image space without requiring user-provided 3D scene representations.
Early GAN-based methods learned shadow synthesis from object masks and image appearance~\citep{zhang2019shadowgan,Liu2020ARShadowGANSG}.
\citet{hong2022shadow} introduced a two-stage model that predicts a shadow mask before refining its appearance using an image-formation model.
SGDiffusion~\citep{liu2024shadow} uses the composite image and object mask to condition a ControlNet-based diffusion model~\citep{zhang2023adding}, while GPSDiffusion~\citep{zhao2025shadow} introduces 2D geometry priors through a predicted rotated shadow bounding box and retrieved shadow shapes.
However, because these methods encode physical constraints only indirectly through learned image-space features and priors, shadow shape and placement can remain unreliable.

\noindent \textbf{Depth and lighting estimation.} Recent progress in monocular geometry estimation makes it feasible to recover useful approximate scene structure from a single image. Monocular depth models perform well on diverse scenes, with \citet{yang2024depthanything,yang2024depth} providing strong relative depth, and \citet{yin2023metric3d,piccinelli2024unidepth} demonstrating accurate metric depth. In this work, we use the off-the-shelf MoGe-2~\citep{wang2025moge2} to recover metric point maps. Illumination can be represented in several forms, including environment maps~\citep{somanath2021hdr,verbin2024eclipse}, light probes~\citep{bai2023local, phongthawee2024diffusionlight}, and parameterized lighting directions~\citep{zhang2019all,dastjerdi2023everlight}. For outdoor shadow generation, we use a parameterized light direction as a compact cue for geometric shadow estimation.

\section{Method}
\label{sec:method}

\begin{figure}[!t]
    \centering
    \includegraphics[width=1.0\linewidth]{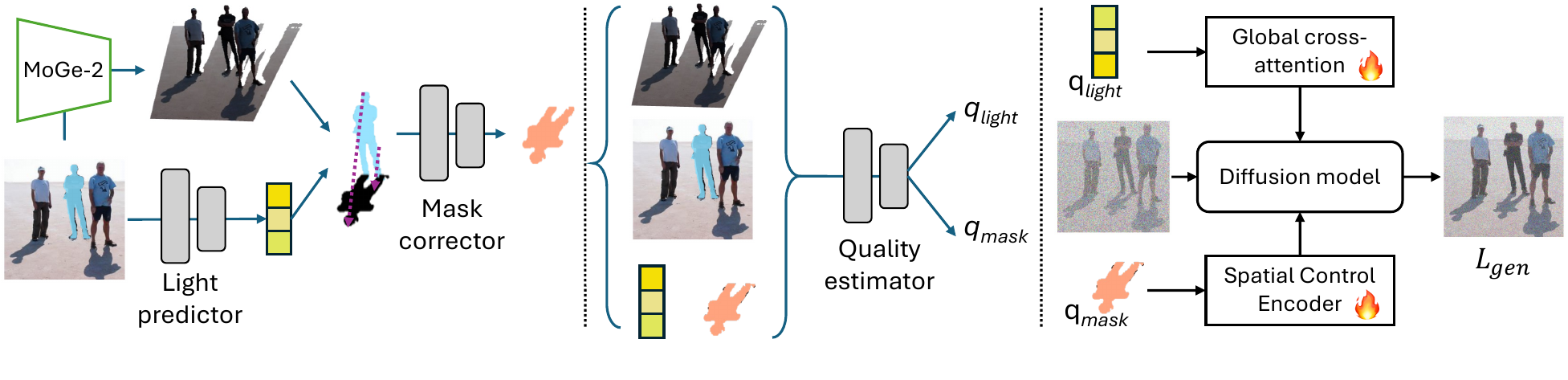}
    \makebox[0.36\linewidth]{(a) Shadow Mask Estimation}%
    \makebox[0.32\linewidth]{(b) Quality Estimation}%
    \makebox[0.32\linewidth]{(c) Shadow Generation}%
    \caption{\textbf{Framework overview.}
    \textbf{(a) Shadow mask estimation.}
    A light predictor estimates a dominant 3D light direction from the input image, object mask, and 3D point map. Together with the point map and object mask, this direction is used to construct a coarse physics-guided soft shadow prior through 3D ray-direction alignment. A mask corrector refines the prior into a full-resolution shadow mask.
    \textbf{(b) Quality estimation.}
    A quality estimator predicts confidence scores $q_{\text{light}}$ and $q_{\text{mask}}$ for the predicted light direction and shadow mask, respectively.
    \textbf{(c) Shadow generation.}
    A frozen diffusion backbone is augmented with two trainable conditioning pathways. The confidence-weighted shadow mask provides dense spatial control, while the predicted light direction and its confidence modulate scene features injected through cross-attention for scene-level lighting guidance.}
    \label{fig:framework}
\end{figure}

Given a composite image $I$ containing a foreground object whose cast shadow is absent, together with its mask $M_{fo}$, our goal is to synthesize a physically plausible cast shadow for the object. 
We use an off-the-shelf monocular geometry estimator~\citep{wang2025moge2} that provides a dense point map $\mathbf{P}$, and we define the hint stack $\mathbf{H} = \mathrm{concat}(I, M_{fo}, \mathbf{P})$ as the base input to our framework.
Our framework, as depicted in Figure~\ref{fig:framework}, has three stages:
\begin{itemize}[leftmargin=*]
\item First, a light predictor estimates a dominant 3D light direction $\hat{\mathbf{l}}$ from $\mathbf{H}$ under the directional-light assumption for outdoor scenes. Together with $\mathbf{P}$ and $M_{fo}$, the predicted direction is used to construct a coarse physics-guided shadow prior through 3D ray-direction alignment. A mask corrector then refines this prior into a full-resolution mask $\tilde{M}_{fs}$.
\item Second, a quality estimator predicts two scalar confidence scores $q_{\text{light}}$ and $q_{\text{mask}}$, measuring the reliability of $\hat{\mathbf{l}}$ and $\tilde{M}_{fs}$, respectively.
\item Third, a frozen Stable Diffusion backbone is augmented with two trainable conditioning pathways: $\tilde{M}_{fs}$, scaled by $q_{\text{mask}}$, conditions a spatial control branch for dense local guidance, while scene features modulated by $\hat{\mathbf{l}}$ and gated by $q_{\text{light}}$ are encoded as cross-attention tokens for global scene-level lighting guidance.
An empty text prompt is used. 
\end{itemize}

\subsection{Light Predictor}
\label{sec:lightpredictor}
We estimate a dominant scene-level 3D light direction from the hint stack $\mathbf{H}$. A ConvNeXt-Small backbone~\citep{liu2022convnet} processes $\mathbf{H}$, followed by global average pooling and a 2-layer MLP that regresses a vector $\mathbf{l} \in \mathbb{R}^3$, normalized to the unit direction $\hat{\mathbf{l}} = \mathbf{l} / \lVert\mathbf{l}\rVert_2$, pointing toward the light source. Training uses a cosine-similarity loss:
\begin{equation}
    \mathcal{L}_{\text{lit}} = 1 - \hat{\mathbf{l}} \cdot \hat{\mathbf{l}}_{\mathrm{gt}},
\end{equation}
where $\hat{\mathbf{l}}_{\mathrm{gt}}$ is obtained using our offline pseudo-labeling pipeline and used only for training supervision (see Appendix~\ref{sec:lightgt}). The light predictor is trained with the mask corrector and frozen thereafter.

\subsection{Shadow Mask Estimation}
\label{sec:mask}
Given the object mask, point map, and predicted light direction, we predict an auxiliary shadow mask $\tilde{M}_{fs}$ using a geometry-based shadow prior followed by a coarse-to-fine refinement network.

\noindent \textbf{Geometry-based shadow prior.}
A receiver pixel is shadowed if the object blocks the light ray that would otherwise reach it. We approximate this by checking whether any object pixel lies along the shadow-flow direction toward the receiver. Let $\mathcal{O}=\{o\mid M_{fo}(o)=1\}$ denote the object pixels and $\mathcal{R}=\{r\mid M_{fo}(r)=0\}$ denote the candidate receiver pixels. For each $o\in\mathcal{O}$ and $r\in\mathcal{R}$, we compute the unit direction $\hat{\mathbf{v}}$ from the object point to the receiver point and measure its alignment with the shadow flow:
\begin{equation}
    \hat{\mathbf{v}} = \frac{\mathbf{P}(r) - \mathbf{P}(o)}
                           {\lVert \mathbf{P}(r) - \mathbf{P}(o) \rVert_2},
    \qquad
    a = \hat{\mathbf{v}} \cdot \mathbf{D},
\end{equation}
where $\mathbf{D} = -\hat{\mathbf{l}}$ is the shadow flow direction from occluder toward receiver. 
The alignment score $a$ is converted to a soft probability via a sigmoid with temperature $T$ and angular tolerance $\tau$, and each receiver pixel takes the maximum score across all object pixels:
\begin{equation}
    s(o,r) = \sigma\!\left(\frac{a - \cos\tau}{T}\right),
    \qquad
    \mathrm{prior}(r) = \max_{o \in \mathcal{O}}\, s(o,r).
\end{equation}
In practice, we compute this prior at $64\times64$ resolution for efficiency and apply a small Gaussian blur for spatial smoothness.

\noindent \textbf{Coarse-to-fine refinement.}
The soft prior captures approximate location but is too coarse and noisy for fine shadow shape and boundaries. We refine it in two stages. At $64{\times}64$, the soft prior, downsampled object mask, and point map are passed through a lightweight network to produce coarse logits, which are then upsampled to $512{\times}512$ as $L_c$. At full resolution, a residual correction $\Delta$ is predicted from $L_c$, $\mathbf{H}$, and a broadcast light map with each pixel set to $\hat{\mathbf{l}}$, then added in logit space.

The final mask is:
\begin{equation}
    \tilde{M}_{fs} = \sigma\!\left(L_c + \Delta\right).
\end{equation}

The refinement network is trained end-to-end using BCE and Dice 
losses~\citep{sudre2017generalised}:
\begin{equation}
    \mathcal{L}_{\text{mask}} = \mathrm{BCE}(\tilde{M}_{fs},\, \hat{M}_{fs})
    + \lambda_1\mathcal{L}_{\mathrm{Dice}}(\tilde{M}_{fs},\hat{M}_{fs}),
\end{equation}
where $\hat{M}_{fs}$ is the ground-truth shadow mask. This module is trained jointly with the light predictor and frozen during generator training.

\subsection{Quality Estimation}
\label{sec:quality}
Light and mask predictions can be unreliable because single-image illumination is ill-posed and the recovered point map is approximate. We therefore predict confidence scores $q_{\text{light}},\ q_{\text{mask}} \in [0,1]$ to gate their influence: reliable cues guide synthesis, while uncertain ones are down-weighted.

Let $\mathbf{H}_{64} = \mathrm{concat}(I_{64},\, M_{fo,64},\, \mathbf{P}_{64})$ and let $\tilde{M}_{fs,64}$ be the downsampled predicted mask. The quality network takes $\mathbf{H}_{64}$ and $\tilde{M}_{fs,64}$ as spatial inputs, along with $\hat{\mathbf{l}}$ and an intermediate light-predictor feature as lighting inputs. 
A convolutional encoder pools $\mathrm{concat}(\mathbf{H}_{64},\, \tilde{M}_{fs,64})$, and two MLP heads predict $q_{\text{light}}$ and $q_{\text{mask}}$; the light head uses the lighting inputs, while the mask head also uses lightweight mask-geometry cues such as centroid offset, elongation, and disagreement with the soft prior.

Light-direction and mask errors are mapped to soft quality targets. Let
$\theta=\arccos(\hat{\mathbf{l}}\cdot\hat{\mathbf{l}}_{\mathrm{gt}})$
and
$e_{\mathrm{mask}}=1-\operatorname{Dice}(\tilde{M}_{fs},\hat{M}_{fs})$
denote the light and mask errors. We define
\begin{equation}
q_{\mathrm{light}}^{*}=\exp(-\theta/\sigma_{\mathrm{ang}}),
\qquad
q_{\mathrm{mask}}^{*}=\exp(-e_{\mathrm{mask}}/\sigma_{\mathrm{mask}}).
\end{equation}
The training objective combines Smooth-$\ell_1$ regression and pairwise ranking:
\begin{equation}
\mathcal{L}_{q}
=
\sum_{c\in\{\mathrm{light},\mathrm{mask}\}}
\left(
\lambda_{\mathrm{reg}}\mathcal{L}^{\mathrm{reg}}_{c}
+
\lambda_{\mathrm{rank}}\mathcal{L}^{\mathrm{rank}}_{c}
\right).
\end{equation}
The regression loss matches the soft targets, while the ranking loss is applied to pairs whose errors differ by more than a margin $m$, set separately for each cue, and assigns higher confidence to the lower-error prediction. The ranking term is the primary supervision signal. The quality network is trained with the light predictor and mask corrector frozen, then frozen during generator training.

\subsection{Shadow Generator}
\label{sec:gen}
We follow the recent diffusion-based shadow generator~\citep{zhao2025shadow} and inject the predicted shadow mask and light direction through confidence-gated spatial and lighting pathways.

\noindent \textbf{Dense spatial conditioning.}
A ControlNet~\citep{zhang2023adding} branch receives the hint stack augmented with the predicted shadow mask. Before concatenation, the mask is scaled by $q_{\text{mask}}$, so unreliable masks contribute weakly while the other hint channels remain unchanged.

\noindent \textbf{Scene-level lighting conditioning.}
The light direction provides a global cue for shadow orientation. A lightweight encoder maps the image and object mask to features, which are modulated by $\hat{\mathbf{l}}$ via FiLM~\citep{perez2018film} with strength gated by $q_{\text{light}}$. The features are projected into tokens and injected into the denoising U-Net via IP-Adapter~\citep{ye2023ip} cross-attention.

\noindent \textbf{Training.} 
Given timestep $t$ and Gaussian noise $\boldsymbol{\epsilon} \sim \mathcal{N}(\mathbf{0}, \mathbf{I})$, the noised latent is:
\begin{equation}
    \mathbf{z}_t = \sqrt{\bar\alpha_t}\,\mathbf{z}_0
    + \sqrt{1 - \bar\alpha_t}\,\boldsymbol{\epsilon},
\end{equation}
and the denoising network is trained to predict the added noise:
\begin{equation}
    \mathcal{L}_{\text{gen}} =
    \mathbb{E}\!\left[\lVert\boldsymbol{\epsilon} -
    f(\mathbf{z}_t, t, \textit{cond})\rVert_2^2\right],
\end{equation}
where $\textit{cond}$ denotes all conditioning inputs. During training, we upweight the noise-prediction loss within a dilated ground-truth shadow region. The denoising U-Net is frozen; only the ControlNet branch and the lighting-conditioning pathway, including its encoder, FiLM layers, and cross-attention adapter, are updated during generator training.

\section{Experiments and Results}

\noindent \textbf{Dataset.}
We conduct experiments on DESOBAV2~\citep{liu2024shadow}, which contains 21,575 scenes with 28,573 image tuples. The public release is split into 27,823 training tuples and 750 test tuples. The test set covers two conditions: background-object-shadow (BOS), where reference background-object–shadow pairs exist in the image; and BOS-free, where there is only a single object–shadow pair. We obtain per-image point maps from MoGe-2~\citep{wang2025moge2} and derive 22,364 approximate light-direction pseudo-labels using our offline pipeline. Light losses use only training tuples with valid directions, while all training tuples are retained for mask and generator training.

\noindent \textbf{Baselines and Metrics.}
We compare against four GAN-based methods, ShadowGAN~\citep{zhang2019shadowgan}, MaskShadowGAN~\citep{hu_iccv2019mask}, ARShadowGAN~\citep{Liu2020ARShadowGANSG}, and SGRNet~\citep{hong2022shadow}, pretrained on DESOBA~\citep{hong2022shadow}. 
For SGDiffusion~\citep{liu2024shadow} and GPSDiffusion~\citep{zhao2025shadow}, we re-evaluate their released checkpoints on the public DESOBAV2 test split using identical preprocessing.
Following prior work~\citep{zhao2025shadow}, we report RMSE and SSIM for image quality, and BER for shadow mask accuracy, evaluated on both the global image (G) and the local foreground-shadow region (L). 
For all diffusion-based methods, we generate five samples per test case and select the sample with the highest local SSIM against the ground truth.

\noindent \textbf{Implementation Details.}
We train the shadow generator with AdamW~\citep{loshchilov2017adamw} at a learning rate of $1\times10^{-5}$ for 30 epochs on two NVIDIA RTX A6000 GPUs with a batch size of 4. Inference uses 50-step DDIM~\citep{song2020denoising}, followed by the post-processing network of~\citet{liu2024shadow}; remaining settings are provided in Appendix~\ref{sec:impl}.

\begin{table}[!t]
\centering
\caption{Comparison with state-of-the-art methods on DESOBAV2 under BOS and BOS-free settings. GR/LR, GS/LS, and GB/LB denote global/local RMSE, SSIM, and BER, respectively. GAN-based methods are pretrained on DESOBA. Diffusion methods use best-of-five selection; \textit{italics} denote single-sample results averaged over five independent draws. Best scores are in \textbf{bold}.}
\label{tab: all test comp}

\small
\setlength{\tabcolsep}{3.0pt}
\renewcommand{\arraystretch}{1.08}

\resizebox{\linewidth}{!}{%
\begin{tabular}{
l
S[table-format=2.3]
S[table-format=2.3]
S[table-format=1.3]
S[table-format=1.3]
S[table-format=1.3]
S[table-format=1.3]
@{\hspace{5pt}}
S[table-format=2.3]
S[table-format=2.3]
S[table-format=1.3]
S[table-format=1.3]
S[table-format=1.3]
S[table-format=1.3]
}
\toprule
\multirow{2}{*}{\textbf{Method}} &
\multicolumn{6}{c}{\textbf{BOS}} &
\multicolumn{6}{c}{\textbf{BOS-free}} \\
\cmidrule(lr){2-7}\cmidrule(lr){8-13}
& {GR$\downarrow$}
& {LR$\downarrow$}
& {GS$\uparrow$}
& {LS$\uparrow$}
& {GB$\downarrow$}
& {LB$\downarrow$}
& {GR$\downarrow$}
& {LR$\downarrow$}
& {GS$\uparrow$}
& {LS$\uparrow$}
& {GB$\downarrow$}
& {LB$\downarrow$} \\
\midrule
ShadowGAN~\citep{zhang2019shadowgan}
& 8.681 & 70.459 & 0.961 & 0.174 & 0.470 & 0.938
& 19.146 & 87.149 & 0.903 & 0.052 & 0.483 & 0.961 \\

Mask-SG~\citep{hu_iccv2019mask}
& 10.450 & 73.776 & 0.938 & 0.186 & 0.485 & 0.966
& 19.662 & 89.637 & 0.895 & 0.054 & 0.488 & 0.972 \\

AR-SG~\citep{Liu2020ARShadowGANSG}
& 8.873 & 69.336 & 0.957 & 0.190 & 0.463 & 0.922
& 19.594 & 84.939 & 0.896 & 0.057 & 0.468 & 0.925 \\

SGRNet~\citep{hong2022shadow}
& 9.017 & 71.582 & 0.961 & 0.189 & 0.446 & 0.887
& 20.883 & 85.841 & 0.894 & 0.056 & 0.450 & 0.881 \\
\midrule
SGDiffusion~\citep{liu2024shadow}
& 7.366 & 51.830 & 0.962 & 0.332 & 0.218 & 0.433
& 14.664 & 54.931 & 0.914 & 0.168 & 0.179 & 0.348 \\

GPSDiffusion~\citep{zhao2025shadow}
& 6.694 & 41.266 & 0.967 & 0.449 & 0.140 & 0.274
& 18.207 & 52.602 & 0.906 & 0.178 & 0.147 & 0.271 \\

\textbf{Ours}
& {\bfseries 4.493} & {\bfseries 33.530}
& {\bfseries 0.974} & {\bfseries 0.533}
& {\bfseries 0.107} & {\bfseries 0.211}
& {\bfseries 9.785} & {\bfseries 36.126}
& {\bfseries 0.934} & {\bfseries 0.322}
& {\bfseries 0.079} & {\bfseries 0.151} \\
\midrule
\textit{GPSDiffusion (average of 5)}
& 7.980 & 50.290 & 0.964 & 0.318 & 0.197 & 0.387
& 21.560 & 61.920 & 0.897 & 0.111 & 0.202 & 0.376 \\

\textit{Ours (average of 5)}
& 5.230 & 39.910 & 0.972 & 0.428 & 0.143 & 0.283
& 11.780 & 43.330 & 0.927 & 0.241 & 0.113 & 0.217 \\
\bottomrule
\end{tabular}%
}
\end{table}

\begin{figure}[!t]
\centering
\includegraphics[width=\linewidth]{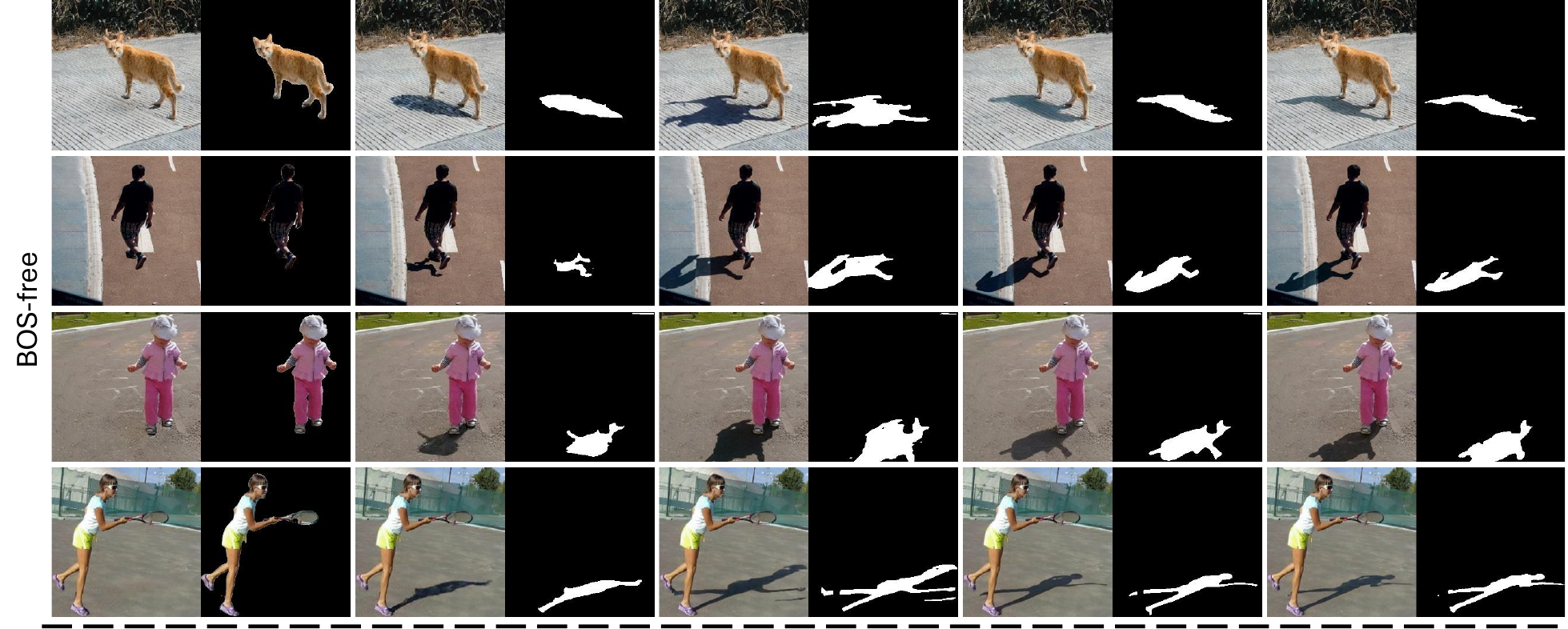}
\includegraphics[width=\linewidth]{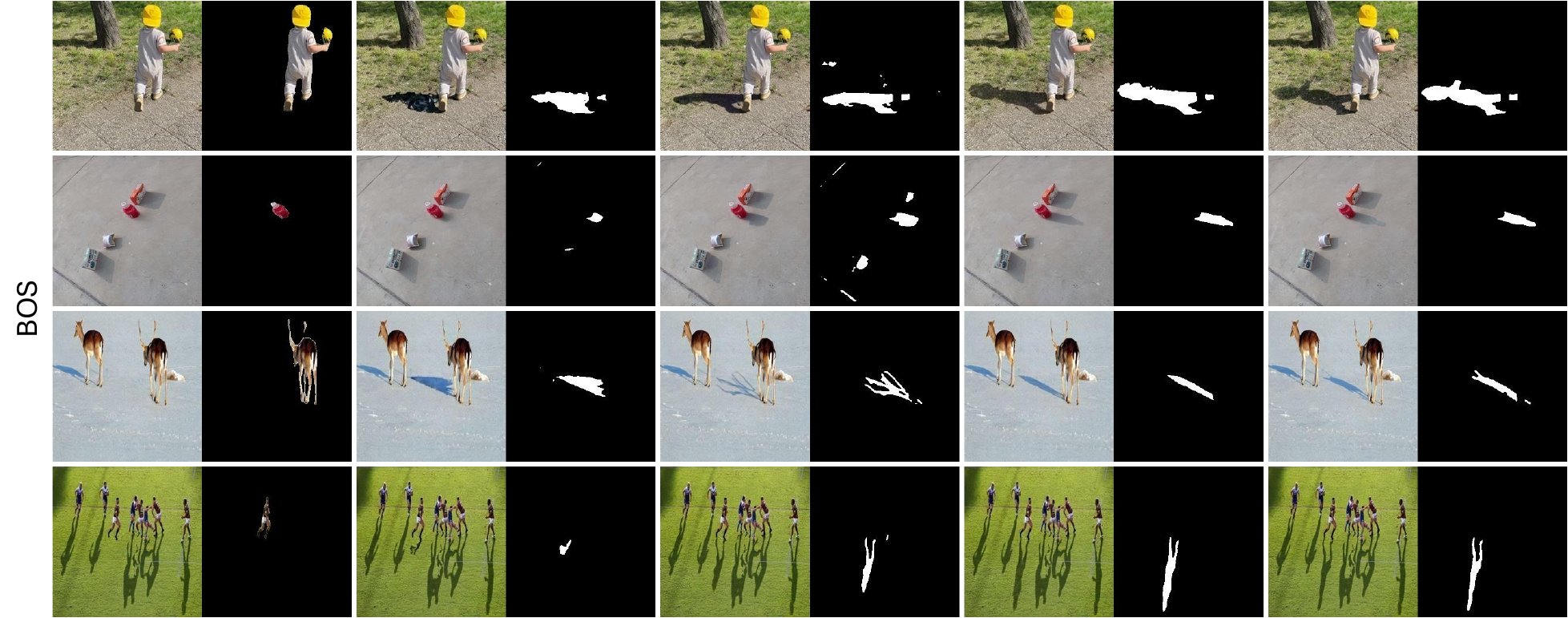}

{\small
  \noindent
  \makebox[0.03\linewidth][c]{}%
  \makebox[0.19\linewidth][c]{Img. \& Obj.}%
  \makebox[0.19\linewidth][c]{SGDiffusion}%
  \makebox[0.19\linewidth][c]{GPSDiffusion}%
  \makebox[0.21\linewidth][c]{Ours}%
  \makebox[0.19\linewidth][c]{GT Img. + Mask}%
}
\caption{\textbf{Qualitative comparison with state-of-the-art methods.}
Results in both BOS (with reference background object--shadow pairs) and BOS-free (with a single object--shadow pair) settings. We compare generated images and predicted shadow masks from SGDiffusion~\citep{liu2024shadow}, GPSDiffusion~\citep{zhao2025shadow}, and our method with the ground truth. Our method produces more accurately localized shadow masks and visually coherent shadows across both settings.}
\label{fig: qualsota}
\end{figure}

\subsection{Comparisons with SOTA}

\noindent \textbf{Quantitative Results.}
Our method achieves the best performance across all metrics in both BOS and BOS-free settings (Table~\ref{tab: all test comp}). Under the best-of-five protocol, compared with GPSDiffusion~\citep{zhao2025shadow}, our method reduces LRMSE by 31\% in BOS-free ($52.602 \rightarrow 36.126$) and 19\% in BOS ($41.266 \rightarrow 33.530$), with consistent global and local SSIM gains. GBER/LBER decrease by 24\%/23\% in BOS and 46\%/44\% in BOS-free.

For single-sample evaluation, we compare the methods using one sample per test case in each of five draws and average the results, without selecting the best sample. Under this protocol, our method outperforms GPSDiffusion across every metric, reducing LRMSE/LBER by 21\%/27\% in BOS and 30\%/42\% in BOS-free. These results show that the gains persist under single-sample inference and do not rely on ground-truth-based best-of-five selection. Across both protocols, the improvements in LRMSE and LBER indicate greater shadow-region fidelity and more accurate localization.

\noindent \textbf{Qualitative Results.}
Figure~\ref{fig: qualsota} compares generated images and predicted shadow masks from SGDiffusion~\citep{liu2024shadow}, GPSDiffusion~\citep{zhao2025shadow}, and our method. In the shown cases, SGDiffusion produces incomplete shadow shapes or inconsistent intensity (rows 4, 6, and 8), while GPSDiffusion can drift or produce inaccurate shadow extent and shape (rows 1, 3, and 7). In contrast, our method anchors shadow location and shape using a physics-guided shadow estimate and dominant lighting cue, producing plausible results on varied receivers, including uneven and textured surfaces (rows 2 and 5).
Additional qualitative comparisons are provided in Appendix~\ref{sec:morequal}.

\begin{figure}[!t]
  \centering
  \includegraphics[width=\linewidth]{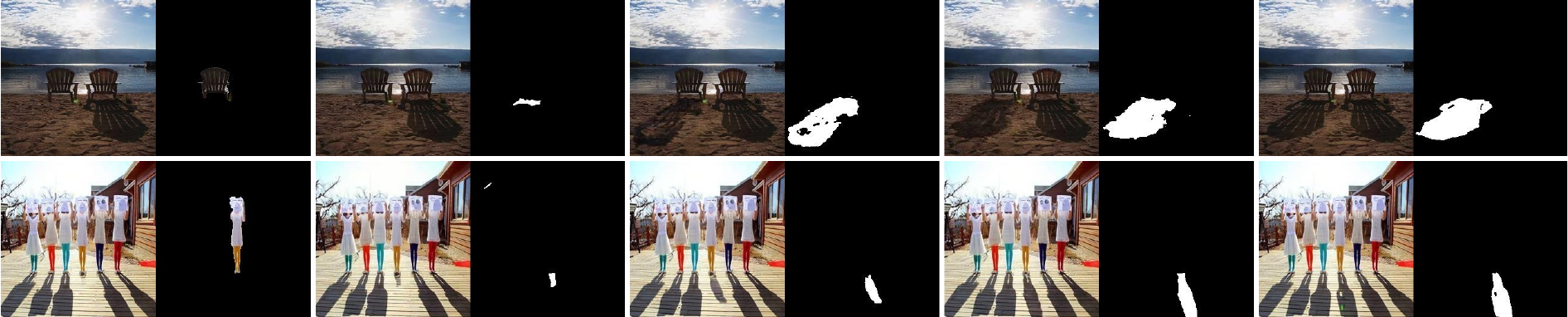}
  {\small
    \makebox[0.19\linewidth][c]{Img. \& Obj.}%
    \makebox[0.19\linewidth][c]{SGDiffusion}%
    \makebox[0.21\linewidth][c]{GPSDiffusion}%
    \makebox[0.19\linewidth][c]{Ours}%
    \makebox[0.19\linewidth][c]{GT Img. + Mask}%
  }
  \caption{\textbf{Challenging cases.}
  (\emph{Top}) Spatially varying illumination makes a single dominant light
  direction ambiguous. (\emph{Bottom}) An out-of-frame shadow truncates the
  reference object--shadow pair.}
  \label{fig: hard}
\end{figure}

\noindent \textbf{Challenging Cases.} 
Reliable light-direction pseudo-labels cannot always be recovered from object--shadow pairs, particularly under spatially varying illumination or out-of-frame shadow truncation. Such cases are excluded from light-predictor supervision. At inference, the predicted confidence scores regulate cue strength during generation. Figure~\ref{fig: hard} shows that our method produces coherent, well-placed shadows, while the baselines produce incomplete or displaced shadows.

\subsection{Performance of the Light and Mask Predictors}

Table~\ref{tab: predictors} summarizes predictor performance on DESOBAV2. We report light-direction angular error on samples with valid light directions and full-resolution shadow-mask Dice on all test samples. 
Light prediction is more accurate in BOS than BOS-free, with mean angular errors of $6.78^\circ$ and $10.11^\circ$, respectively. This gap is consistent with the richer scene-level evidence in BOS, where reference background object--shadow pairs constrain illumination. Without such pairs, light-direction estimation is more ambiguous. Mask quality remains similar across the two settings, with mean Dice scores of $0.502$ and $0.486$. Overall, the predictors provide useful but imperfect cues, motivating confidence-aware conditioning rather than treating their outputs as fixed constraints.
See Appendices~\ref{sec:light_vis} and~\ref{sec:mask_vis} for predictor visualizations.

\begin{table}[!t]
\centering
\caption{Predictor statistics on the DESOBAV2 test set. $N$ denotes the number of evaluated samples. Light-direction error is reported in degrees on samples with valid directions, while mask quality is reported as full-resolution Dice on all samples.}
\label{tab: predictors}

\setlength{\tabcolsep}{4.0pt}
\renewcommand{\arraystretch}{1.08}

\begin{minipage}[t]{0.48\linewidth}
\centering
\begin{tabular*}{\linewidth}{@{\extracolsep{\fill}} l c c c}
\toprule
\textbf{Light} & \textbf{N} & \textbf{Mean} & \textbf{p50 / p90} \\
\midrule
BOS      & 375 & 6.780  & 5.90 / 11.40 \\
BOS-free & 235 & 10.110 & 7.86 / 20.94 \\
\bottomrule
\end{tabular*}
\end{minipage}
\hfill
\begin{minipage}[t]{0.48\linewidth}
\centering
\begin{tabular*}{\linewidth}{@{\extracolsep{\fill}} l c c c}
\toprule
\textbf{Mask} & \textbf{N} & \textbf{Mean} & \textbf{p50 / p90} \\
\midrule
BOS      & 500 & 0.502 & 0.534 / 0.712 \\
BOS-free & 250 & 0.486 & 0.513 / 0.649 \\
\bottomrule
\end{tabular*}
\end{minipage}
\end{table}

\subsection{Performance of the Quality Estimation Network}
\label{sec: qualitynet}

The quality network predicts $q_{\text{mask}}, q_{\text{light}} \in [0,1]$ to represent the relative reliability of the shadow-mask and light-direction cues. Table~\ref{tab: quality_stats} evaluates whether these scores track their corresponding prediction errors. We report Pearson correlation with error and the mean errors of the highest- and lowest-scoring 20\% of samples. For both cues and both test settings, the correlations are negative, and the highest-scoring samples have substantially lower errors than the lowest-scoring samples. Thus, the predicted scores provide meaningful reliability signals for confidence-aware conditioning. Beyond tracking cue error, the predicted scores outperform fixed $q_{\mathrm{mask}}=q_{\mathrm{light}}\in\{0,0.5,1\}$ on all six BOS-free metrics, supporting the benefit of input-adaptive conditioning (Appendix~\ref{sec:qsweep}).

\begin{figure}[!t]
   \centering
   \includegraphics[width=\linewidth]{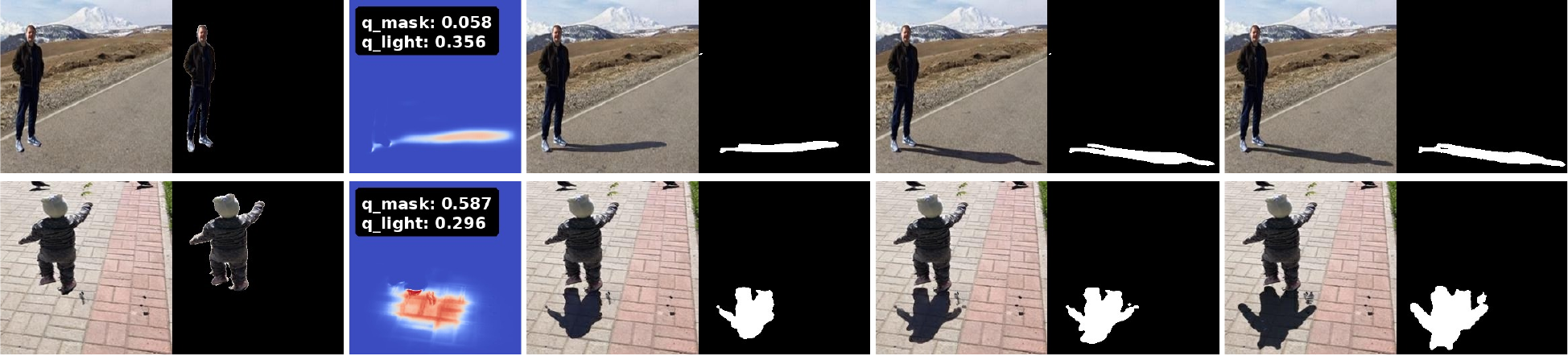}
   {\small
      \makebox[0.22\linewidth]{Img. \& Obj. }%
      \makebox[0.12\linewidth]{Shadow Est.}%
      \makebox[0.22\linewidth]{Forced $q=1$}%
      \makebox[0.22\linewidth]{Predicted $q$}%
      \makebox[0.22\linewidth]{GT }%
    }
    \caption{\textbf{Quality-guided generation.}
    We compare forcing $q_{\text{mask}}=q_{\text{light}}=1$ with using their predicted values. Full-strength conditioning can propagate errors from inaccurate cues, whereas the predicted scores attenuate unreliable guidance and produce better-aligned shadows and masks.}
   \label{fig: qualper}
\end{figure}

Figure~\ref{fig: qualper} further illustrates how the quality estimates affect generation when the predicted conditioning cues are inaccurate. Setting $q_{\text{mask}}=q_{\text{light}}=1$ applies both cues at full strength and can propagate their errors. In contrast, the predicted scores attenuate unreliable conditioning, allowing the generator to rely more on learned image priors and yield better-aligned shadows and masks.

\begin{table}[!t]
\centering
\caption{Reliability of predicted quality scores on DESOBAV2 test set. We report Pearson correlation with the corresponding cue error and mean errors among the highest- and lowest-scoring 20\% of samples. Mask error is $1-\operatorname{Dice}$, and light error is angular error in degrees. Negative correlations and lower top-group errors indicate reliable ranking.}
\label{tab: quality_stats}

\setlength{\tabcolsep}{6.0pt}
\renewcommand{\arraystretch}{1.08}

\begin{tabular}{l l c c c c}
\toprule
\textbf{Score} & \textbf{Split} & \textbf{N} &
\textbf{Pearson $r$} & \textbf{Top 20\% Error} &
\textbf{Bottom 20\% Error} \\
\midrule
$q_{\text{mask}}$
& BOS      & 500 & -0.751 & 0.368 & 0.743 \\
& BOS-free & 250 & -0.556 & 0.430 & 0.649 \\
\midrule
$q_{\text{light}}$
& BOS      & 375 & -0.355 & 4.583 & 10.461 \\
& BOS-free & 235 & -0.518 & 6.308 & 16.240 \\
\bottomrule
\end{tabular}
\end{table}

\subsection{Ablation Studies}

Table~\ref{tab: ablation} evaluates point-map guidance, predictor-based conditioning, and quality-aware conditioning on DESOBAV2 (BOS-free). The point-map-only configuration substantially outperforms GPSDiffusion (Configuration~2 vs.\ 1), supporting the value of monocular geometry for shadow placement. Applying jointly trained light and mask predictors without quality gating slightly degrades all metrics (Configuration~3 vs.\ 2), indicating that direct conditioning on imperfect predictions can be counterproductive. Quality-gated light conditioning alone improves global RMSE and SSIM but remains weaker in BER (Configuration~4), showing that the global lighting cue cannot replace dense mask guidance for accurate localization. The full configuration achieves the best overall performance, supporting the combined use of geometric, lighting, mask, and quality-aware guidance.

\begin{figure}[!t]
   \centering
   \includegraphics[width=\linewidth]{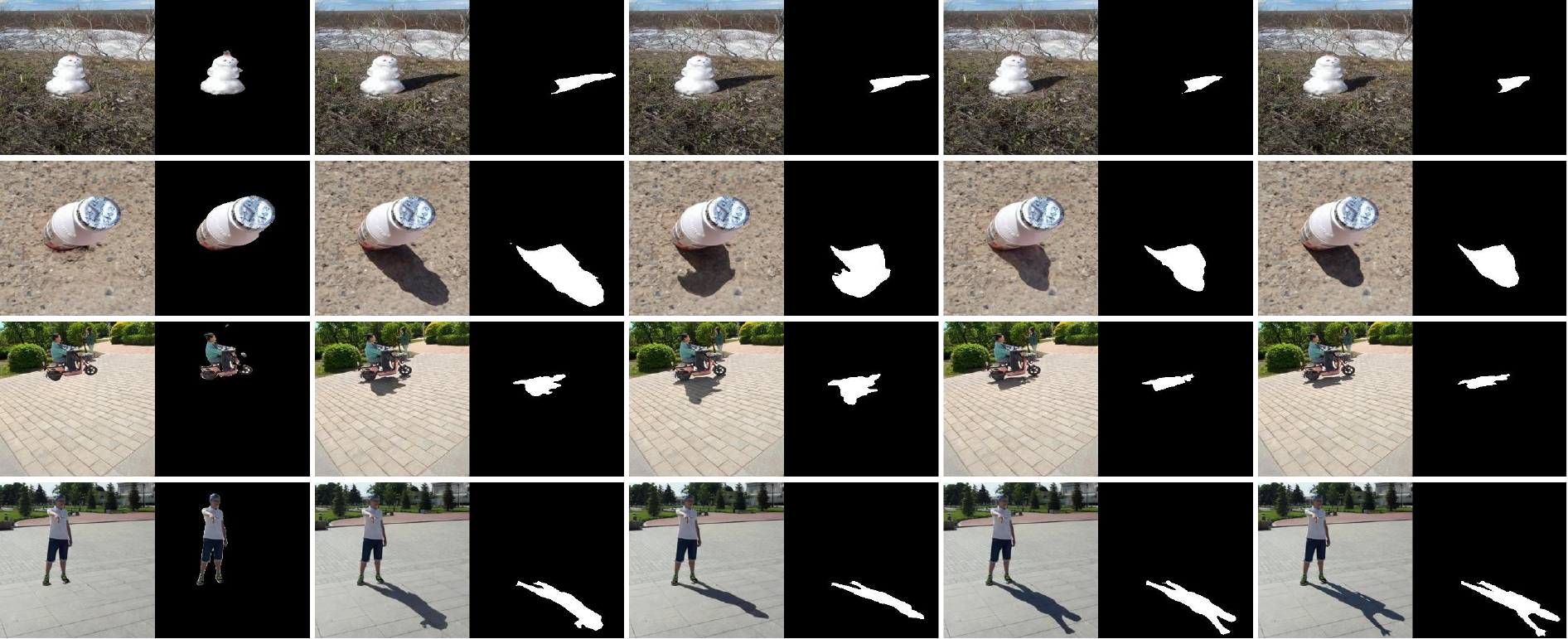}
   {\small
      \makebox[0.19\linewidth]{Img. \& Obj. }%
      \makebox[0.19\linewidth]{PointMap (PM)}%
      \makebox[0.22\linewidth]{PM+Predictors}%
      \makebox[0.19\linewidth]{Ours}%
      \makebox[0.19\linewidth]{GT }%
    }
   \caption{\textbf{Ablations.}
   Point-map guidance provides coarse shadow placement, while ungated predictor conditioning can introduce errors in shadow extent or direction. Our full model modulates these cues using their predicted quality, improving shadow shape and localization.
   }
   \label{fig: ablation}
\end{figure}

Figure~\ref{fig: ablation} visualizes this progression. Point-map guidance provides coarse shadow placement, while ungated predictor conditioning can introduce errors in shadow extent or direction. The full model modulates these cues using their predicted quality, improving shadow shape and localization.

\begin{table}[!t]
\centering
\caption{Ablation of the predictor and quality-network design on the DESOBAV2 BOS-free setting. Configuration~1 is GPSDiffusion. $\dagger$ marks predictors updated during generator training; Configuration~3 applies their outputs without quality gating. Best displayed scores are in \textbf{bold}.}
\label{tab: ablation}

\setlength{\tabcolsep}{5.0pt}
\renewcommand{\arraystretch}{1.08}

\resizebox{\linewidth}{!}{%
\begin{tabular}{l cccc | cccccc}
\toprule
\textbf{Config.} &
\textbf{PointMap} &
\textbf{Light} &
\textbf{Mask} &
\textbf{Quality} &
\textbf{GRMSE}$\downarrow$ &
\textbf{LRMSE}$\downarrow$ &
\textbf{GSSIM}$\uparrow$ &
\textbf{LSSIM}$\uparrow$ &
\textbf{GBER}$\downarrow$ &
\textbf{LBER}$\downarrow$ \\
\midrule
1 & -- & -- & -- & --
& 18.207 & 52.602 & 0.906 & 0.178 & 0.147 & 0.271 \\

2 & \checkmark & -- & -- & --
& 10.041 & 37.010 & 0.933 & 0.317 & 0.082 & 0.157 \\

3 & \checkmark & $\dagger$ & $\dagger$ & --
& 10.295 & 37.976 & 0.932 & 0.316 & 0.089 & 0.170 \\

4 & \checkmark & \checkmark & -- & \checkmark
& 9.873 & 37.651 & \textbf{0.934} & 0.319 & 0.087 & 0.168 \\

5 & \checkmark & \checkmark & \checkmark & \checkmark
& \textbf{9.785} & \textbf{36.126}
& \textbf{0.934} & \textbf{0.322}
& \textbf{0.079} & \textbf{0.151} \\
\bottomrule
\end{tabular}%
}
\end{table}

\section{Conclusion}
We present a physics-guided framework for generating cast shadows for foreground objects in outdoor composite images. Our framework explicitly links approximate monocular scene geometry and a predicted dominant light direction to cast-shadow placement through a coarse geometric estimate. Rather than treating these uncertain physical estimates as exact constraints, the framework predicts quality scores that regulate the influence of the resulting shadow-mask and lighting cues during diffusion refinement. This combination provides an explicit spatial anchor while retaining learned flexibility for realistic shadow appearance. Experiments on DESOBAV2 demonstrate state-of-the-art shadow-region fidelity and localization.

\subsection*{AI use statement}

We used generative AI tools to assist with module implementation, support the interpretation and presentation of experimental results, and edit the manuscript for readability, clarity, and concision.
The authors reviewed all AI-assisted work, tested the implementations, verified interpretations and numerical claims against experimental results, and checked manuscript edits for accuracy. The authors take full responsibility for the final content of this work.

\bibliography{references}
\bibliographystyle{iclr2027_conference}

\appendix

% \noindent In this supplementary material, we provide additional details and analyses that complement the main paper:

% \begin{enumerate}
%     \item \textbf{Visualization of intermediate predictions.}
%     \begin{itemize}
%         \item \cref{sec:light_vis}: Visualization of estimated light directions.
%         \item \cref{sec:mask_vis}: Visualization of shadow mask predictions.
%         \item \cref{sec:quality_vis}: Effect of the predicted quality scores on generation.
%     \end{itemize}
%     \item \textbf{Additional implementation details} of the training procedure and model components (\cref{sec:impl}).
%     \item \textbf{Additional ablation study} using oracle conditioning to analyze the upper bound of the framework (\cref{sec:oracle}).
%     \item \textbf{Pseudo light direction acquisition pipeline} (\cref{sec: light}).
%     \item \textbf{More qualitative results} (\cref{sec:morequal}).
% \end{enumerate}

\section{Visualization of Light-Direction Predictions}
\label{sec:light_vis}

\begin{figure}[!t]
  \centering
  \includegraphics[width=\linewidth]{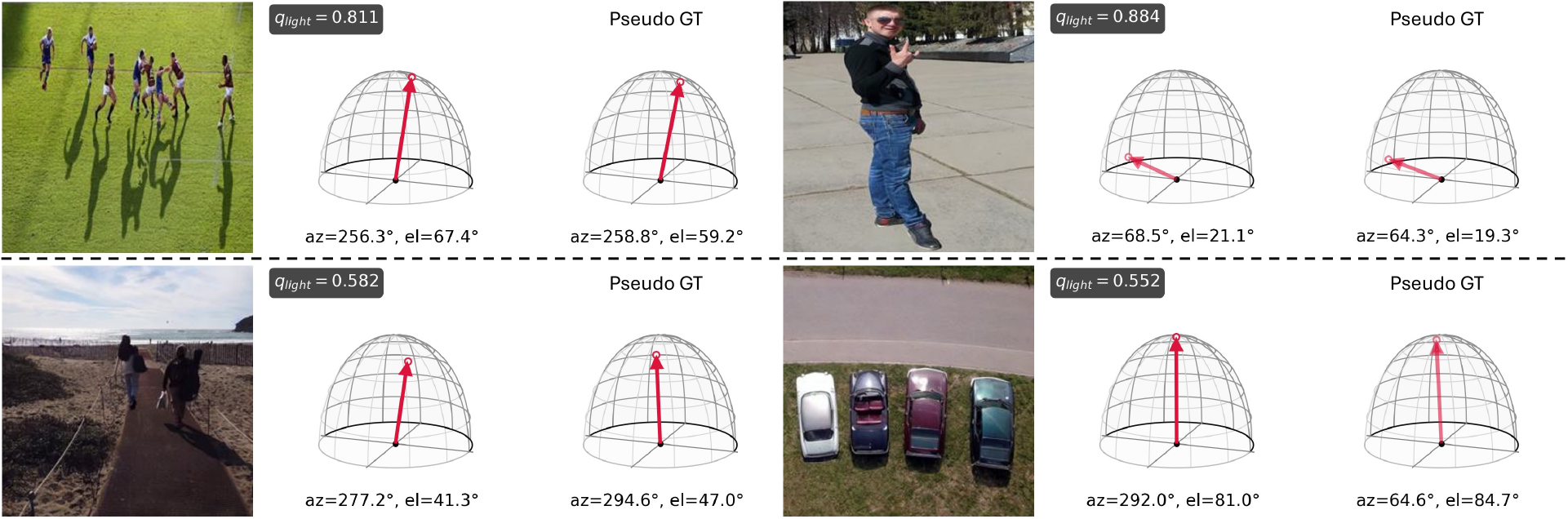}
  \caption{\textbf{Light-direction predictions.}
  Predicted and pseudo-labeled light directions are shown as arrows pointing toward the light source on a hemisphere defined in the input-view coordinate frame. Azimuth and elevation are shown for each direction, together with $q_{\text{light}}$ for the prediction.}
  \label{fig: lightper}
\end{figure}

Figure~\ref{fig: lightper} compares our predicted light directions with the light-direction pseudo-labels produced by the offline labeling pipeline (see Appendix~\ref{sec:lightgt}). We visualize each direction as an arrow on a hemisphere, together with its azimuth and elevation, and report the predicted light-quality score $q_{\text{light}}$. The hemisphere is defined in the input-view coordinate frame and is therefore aligned with the camera viewpoint rather than a world coordinate frame. Accordingly, the directions and angles should be interpreted relative to the image view. For readability, the front half of the hemisphere is rendered more lightly. The visualizations are consistent with the quantitative results in Table~\ref{tab: quality_stats}: predictions closer to their pseudo-labels generally receive higher $q_{\text{light}}$ values, while predictions with larger angular errors receive lower scores.

\section{Visualization of Shadow-Mask Predictions}
\label{sec:mask_vis}

\begin{figure}[!t]
  \centering
  \includegraphics[width=\linewidth]{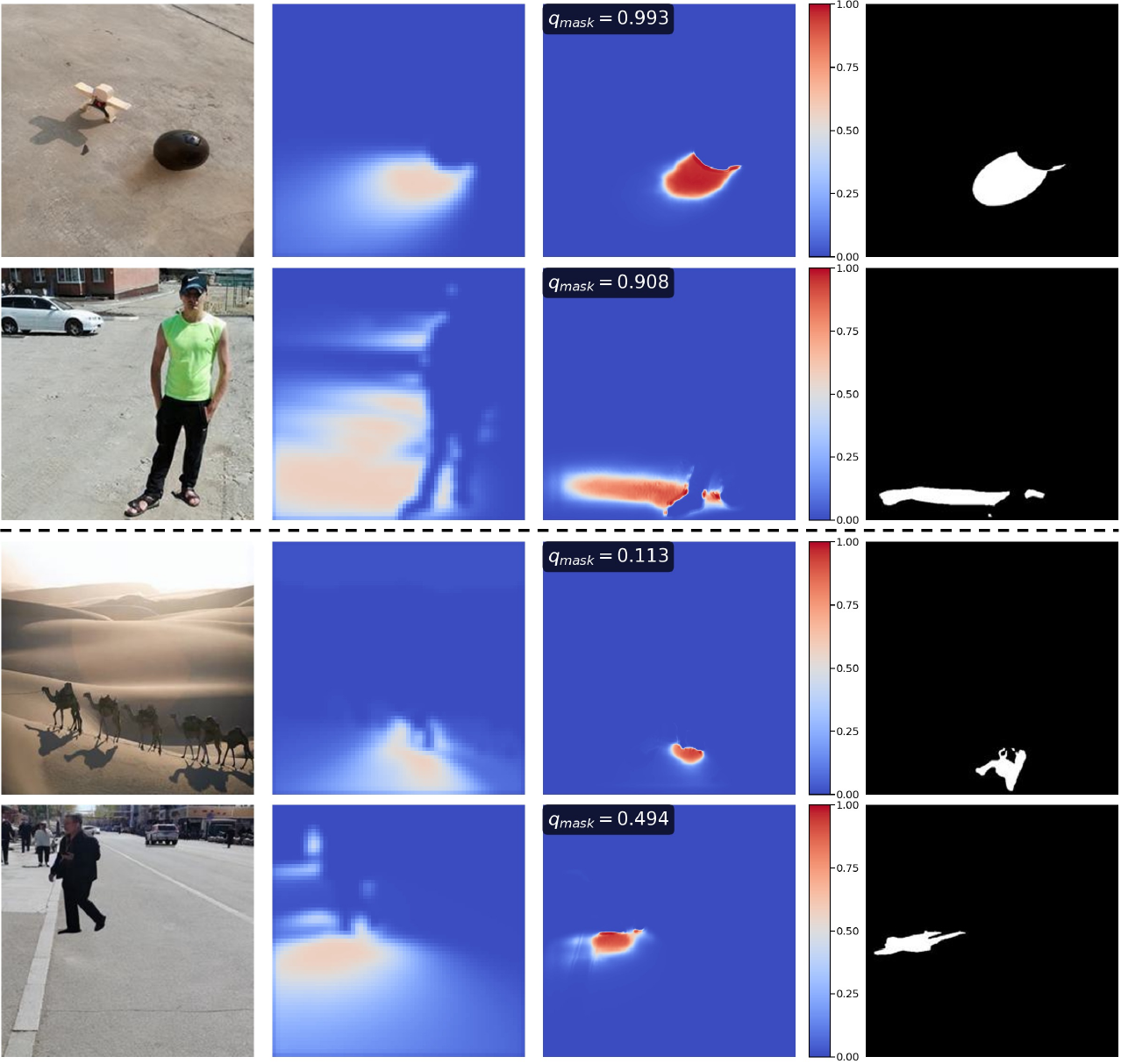}
  \caption{\textbf{Shadow-mask predictions.}
  The geometry-derived soft shadow prior and refined shadow-mask prediction are shown as probability heatmaps, while the ground-truth mask is binary. We also report the scalar quality score $q_{\text{mask}}$ for each refined prediction.}
  \label{fig: maskper}
\end{figure}

Figure~\ref{fig: maskper} compares the geometry-derived soft shadow prior, the refined shadow-mask prediction, and the ground-truth shadow mask. The prior and refined prediction are visualized as probability heatmaps, while the ground-truth mask is binary. The soft prior provides a coarse estimate of shadow orientation and spatial support from the recovered geometry and predicted light direction. The mask predictor refines this estimate to better capture the ground-truth shadow shape and extent. We additionally report the scalar mask-quality score $q_{\text{mask}}$ for each refined prediction. The visualizations follow the trend quantified in Table~\ref{tab: quality_stats}: higher $q_{\text{mask}}$ values generally correspond to better agreement with the ground-truth mask, while lower values are associated with larger mask errors.

\section{Sensitivity to Quality Conditioning}
\label{sec:qsweep}

We further evaluate whether per-input quality prediction is more effective than applying a fixed conditioning strength. At inference, we replace both predicted quality scores with a shared constant $q_{\mathrm{mask}}=q_{\mathrm{light}}\in\{0,0.5,1\}$ while keeping the predicted cues and all model parameters unchanged. Table~\ref{tab:qsweep} reports results on the DESOBAV2 BOS-free setting.

\begin{table}[!t]
\centering
\caption{Sensitivity to quality conditioning on DESOBAV2 BOS-free. Fixed values are applied to both $q_{\mathrm{mask}}$ and $q_{\mathrm{light}}$ at inference. All results use the best-of-five protocol. Best scores are in \textbf{bold}.}
\label{tab:qsweep}

\setlength{\tabcolsep}{5.0pt}
\renewcommand{\arraystretch}{1.08}

\begin{tabular}{l c c c c c c}
\toprule
\textbf{$q$ setting} &
\textbf{GRMSE}$\downarrow$ &
\textbf{LRMSE}$\downarrow$ &
\textbf{GSSIM}$\uparrow$ &
\textbf{LSSIM}$\uparrow$ &
\textbf{GBER}$\downarrow$ &
\textbf{LBER}$\downarrow$ \\
\midrule
Fixed $0$ &
11.00 &
38.84 &
0.9244 &
0.2939 &
0.0803 &
0.1535 \\

Fixed $0.5$ &
10.97 &
38.69 &
0.9246 &
0.2964 &
0.0799 &
0.1528 \\

Fixed $1$ &
11.24 &
39.26 &
0.9240 &
0.2950 &
0.0824 &
0.1573 \\
\midrule
Predicted &
\textbf{9.78} &
\textbf{36.13} &
\textbf{0.9335} &
\textbf{0.3225} &
\textbf{0.0793} &
\textbf{0.1514} \\
\bottomrule
\end{tabular}
\end{table}

The predicted quality scores achieve the best performance across all six metrics. Among the fixed settings, $q=0.5$ generally performs best, but it remains consistently below per-input prediction, while fully trusting both cues with $q=1$ produces the weakest results on five of the six metrics. These results support adapting the conditioning strength to the estimated reliability of each input. Because the fixed values are applied to a model trained with predicted quality scores, this experiment is interpreted as an inference-time sensitivity analysis rather than a comparison with separately trained fixed-gating models.

\section{Additional Ablations with Oracle Conditioning}
\label{sec:oracle}

Table~\ref{tab:gts} extends the main ablation with background-mask and oracle-conditioning variants. In Row~5, the light predictor receives a background mask instead of the foreground object mask. This variant remains competitive but consistently underperforms Row~6, supporting foreground-object conditioning for estimating the relevant light direction.

For a matched oracle comparison, Rows~$6^\ast$ and~7 are evaluated on the 235 BOS-free test samples with valid light-direction pseudo-labels. Row~$6^\ast$ evaluates the full learned model on this subset. In Row~7, the predicted light direction and shadow mask are replaced by the offline pseudo-labeled direction and ground-truth mask, respectively, while the quality estimator is bypassed. Oracle conditioning substantially improves all metrics, showing that the generator can effectively use more accurate conditioning and that improving light and mask estimation offers substantial additional headroom.

\begin{table}[!t]
\centering
\caption{Additional conditioning ablations on the DESOBAV2 BOS-free setting. Rows~1--6 are evaluated on all 250 samples. Row~$6^\ast$ reevaluates the full model on the 235 samples with valid light-direction pseudo-labels, providing a matched comparison with the oracle in Row~7. \texttt{bg} denotes background-mask conditioning, \texttt{PL} denotes the offline pseudo-labeled light direction, and \texttt{GT} denotes the ground-truth shadow mask. Best learned-setting scores on the full test set are in \textbf{bold}. $\dagger$ indicates that the module is not frozen during training.}
\label{tab:gts}

\setlength{\tabcolsep}{5.0pt}
\renewcommand{\arraystretch}{1.08}

\resizebox{0.95\linewidth}{!}{%
\begin{tabular}{l cccc | cccccc}
\toprule
\textbf{Config.} &
\textbf{PointMap} &
\textbf{Light} &
\textbf{Mask} &
\textbf{Quality} &
\textbf{GRMSE}$\downarrow$ &
\textbf{LRMSE}$\downarrow$ &
\textbf{GSSIM}$\uparrow$ &
\textbf{LSSIM}$\uparrow$ &
\textbf{GBER}$\downarrow$ &
\textbf{LBER}$\downarrow$ \\
\midrule
1 & -- & -- & -- & -- &
18.207 & 52.602 & 0.906 & 0.178 & 0.147 & 0.271 \\

2 & \checkmark & -- & -- & -- &
10.041 & 37.010 & 0.933 & 0.317 & 0.082 & 0.157 \\

3 & \checkmark & $\dagger$ & $\dagger$ & -- &
10.295 & 37.976 & 0.932 & 0.316 & 0.089 & 0.170 \\

4 & \checkmark & \checkmark & -- & \checkmark &
9.873 & 37.651 & \textbf{0.934} & 0.319 & 0.087 & 0.168 \\

5 & \checkmark & \texttt{bg} & \checkmark & \checkmark &
9.916 & 36.996 & 0.933 & 0.316 & 0.083 & 0.160 \\

6 & \checkmark & \checkmark & \checkmark & \checkmark &
\textbf{9.785} & \textbf{36.126} & \textbf{0.934} &
\textbf{0.322} & \textbf{0.079} & \textbf{0.151} \\
\midrule
$6^\ast$ & \checkmark & \checkmark & \checkmark & \checkmark &
9.812 & 36.296 & 0.934 & 0.322 & 0.079 & 0.152 \\

7 & \checkmark & \texttt{PL} & \texttt{GT} & -- &
4.345 & 18.362 & 0.957 & 0.583 & 0.016 & 0.032 \\
\bottomrule
\end{tabular}%
}
\end{table}

\section{Limitations}
\label{sec:limitations}

\begin{figure}[!t]
\centering
\includegraphics[width=\linewidth]{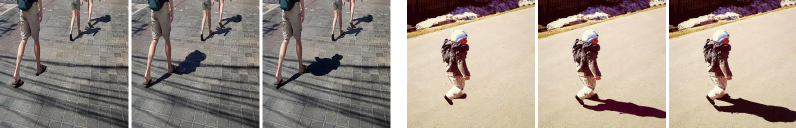}

{\small
\makebox[0.16\linewidth][c]{Input}%
\hspace{0.004\linewidth}%
\makebox[0.16\linewidth][c]{Ours}%
\hspace{0.004\linewidth}%
\makebox[0.16\linewidth][c]{GT}%
\hspace{0.021\linewidth}%
\makebox[0.16\linewidth][c]{Input}%
\hspace{0.004\linewidth}%
\makebox[0.16\linewidth][c]{Ours}%
\hspace{0.004\linewidth}%
\makebox[0.16\linewidth][c]{GT}%
}
\caption{\textbf{Representative limitations.}
Each group shows the input, our result, and ground truth.
(\emph{Left}) Incomplete foreground geometry can lead to an incorrect shadow shape.
(\emph{Right}) Without an explicit attenuation cue, the generated shadow can have plausible placement but inaccurate intensity.}
\label{fig:limitations}
\end{figure}

Figure~\ref{fig:limitations} illustrates two representative limitations of our method.
First, when the foreground object is truncated or occluded, the recovered geometry does not capture its unseen portions. The resulting geometric shadow estimate can therefore be incomplete, leaving the generator to infer the missing shadow structure from learned priors and sometimes producing an incorrect shadow shape.
Second, the dominant light-direction cue constrains shadow orientation but does not explicitly represent attenuation strength. Consequently, the generated shadow can have plausible placement but differ from the ground truth in intensity.
Future work could incorporate object-shape completion and explicit attenuation estimation within the reliability-aware conditioning framework.

\section{Additional Implementation Details}
\label{sec:impl}

This section complements the main paper with additional architectural and optimization details to support reproducibility.

\paragraph{Predictor training and shadow-prior settings.}
The light predictor and mask corrector are jointly optimized for 10 epochs using AdamW with a learning rate of $1\times10^{-4}$, weight decay $1\times10^{-2}$, and batch size 8. The light and mask losses have equal weights, $w_{\mathrm{lit}}=w_{\mathrm{mask}}=1$. For mask refinement, we use
\begin{equation}
\mathcal{L}_{\mathrm{mask}}
=
\operatorname{BCE}
+
0.2\,\mathcal{L}_{\mathrm{Dice}}.
\end{equation}
The geometry-derived soft shadow prior is computed at $64\times64$ resolution with angular tolerance $\tau=10^\circ$ and sigmoid temperature $T=0.05$. We then apply a $5\times5$ Gaussian filter with standard deviation $1.0$ to smooth the prior before mask refinement.

\paragraph{Quality-estimator architecture.}
The refined shadow mask is area-downsampled to $64\times64$, normalized by its mean value, and clipped to $[0,5]$ before being concatenated with $\mathbf{H}_{64}$. The spatial encoder consists of three stride-2 convolutional layers with 32, 64, and 64 output channels. Mean-and-standard-deviation pooling converts the resulting feature map into a 128D descriptor.

The light-quality head concatenates this descriptor with the predicted 3D light direction and the 256D feature from the light predictor. The mask-quality head concatenates the descriptor with the predicted direction and a 4D geometric feature comprising the 2D centroid offset, log elongation, and mean absolute disagreement between the refined mask and soft shadow prior. Both heads use two hidden layers of width 256 followed by a scalar output.

\paragraph{Quality-estimator training.}
The light and mask predictors remain frozen while training the quality estimator. We train it for 10 epochs with AdamW using a learning rate of $1\times10^{-4}$, weight decay $1\times10^{-2}$, and batch size 8. The quality-target scales are
\begin{equation}
\sigma_{\mathrm{ang}}=10^\circ,
\qquad
\sigma_{\mathrm{mask}}=0.70.
\end{equation}
We use $\lambda_{\mathrm{reg}}=0.2$ and $\lambda_{\mathrm{rank}}=1.0$ for both quality heads. The pairwise ranking margins are $m_{\mathrm{light}}=2^\circ$ for angular error and $m_{\mathrm{mask}}=0.05$ for Dice error, with ranking temperature $1.0$. Mask-quality targets and ranking errors are computed after area-downsampling the predicted and ground-truth masks to $64\times64$.

In addition to the objective in the main paper, we use a small auxiliary correlation regularizer:
\begin{equation}
\mathcal{L}_{q}^{\mathrm{total}}
=
\mathcal{L}_{q}
+
\lambda_{\mathrm{area}}
\operatorname{Corr}\!\left(
A(\tilde{M}_{fs,64}), z_{\mathrm{mask}}
\right)^2,
\end{equation}
where $\operatorname{Corr}$ denotes batchwise Pearson correlation, $A(\tilde{M}_{fs,64})$ is the mean predicted-mask area, and $z_{\mathrm{mask}}$ is the pre-sigmoid mask-quality logit. We set $\lambda_{\mathrm{area}}=0.1$.

\paragraph{Generator-specific settings.}
Training uses $512\times512$ inputs and $64\times64$ diffusion latents. To construct the shadow-region loss weights, we resize the ground-truth shadow mask to $64\times64$ and dilate it once using a $6\times6$ kernel. Pixels whose resulting mask value is at least $0.6$ receive denoising-loss weight 10, while all remaining pixels receive weight 1. At inference, classifier-free guidance uses a scale of 9.0 and DDIM sampling uses $\eta=0$.

\paragraph{Computational cost.}
Generator training uses approximately 41 GiB of memory per GPU and takes approximately 80 minutes per epoch; the training server additionally has dual AMD EPYC 7352 CPUs. Under the five-sample, 50-step evaluation protocol, our method requires approximately 50 seconds per test case, compared with 44 seconds for SGDiffusion~\citep{liu2024shadow} and 47 seconds for GPSDiffusion~\citep{zhao2025shadow}. Peak GPU memory is 32.5 GiB for our method, compared with 18.4 GiB and 27.6 GiB for SGDiffusion and GPSDiffusion, respectively.

\section{Light-Direction Pseudo-Label Acquisition}
\label{sec:lightgt}

We estimate approximate dominant light directions offline from the object--shadow pairs available in DESOBAV2. For each scene, MoGe-2~\citep{wang2025moge2} predicts a dense 3D point map $\mathbf{p}(x,y)\in\mathbb{R}^3$. We select one foreground object--shadow pair and use its object mask $M_{fo}$ and observed shadow mask $M_{fs}$ to sample two sets of points: occluder points on the object and receiver points on the shadowed surface. The resulting direction is shared by all tuples from that scene.

\paragraph{Foreground pair selection.}
For each scene, we select the object--shadow pair whose shadow mask has the largest area. If the mask is fragmented, only its largest connected component is retained. Cases with extremely small shadow regions, unreliable geometry, or failed optimization are excluded from light-direction supervision.

\paragraph{Receiver surface estimation.}
We approximate the receiver surface by fitting a plane to 3D points sampled inside the shadow mask. 
Given receiver points $\{\mathbf{r}_k\}$, the plane normal is obtained via PCA on the covariance matrix
\begin{equation}
\mathbf{\Sigma}=\frac{1}{K}\sum_k(\mathbf{r}_k-\bar{\mathbf{r}})(\mathbf{r}_k-\bar{\mathbf{r}})^\top ,
\end{equation}
where $\bar{\mathbf{r}}$ is the centroid. The eigenvector corresponding to the smallest eigenvalue defines the plane normal $\mathbf{n}$.

\begin{figure}[!t]
   \centering
   \includegraphics[width=\linewidth]{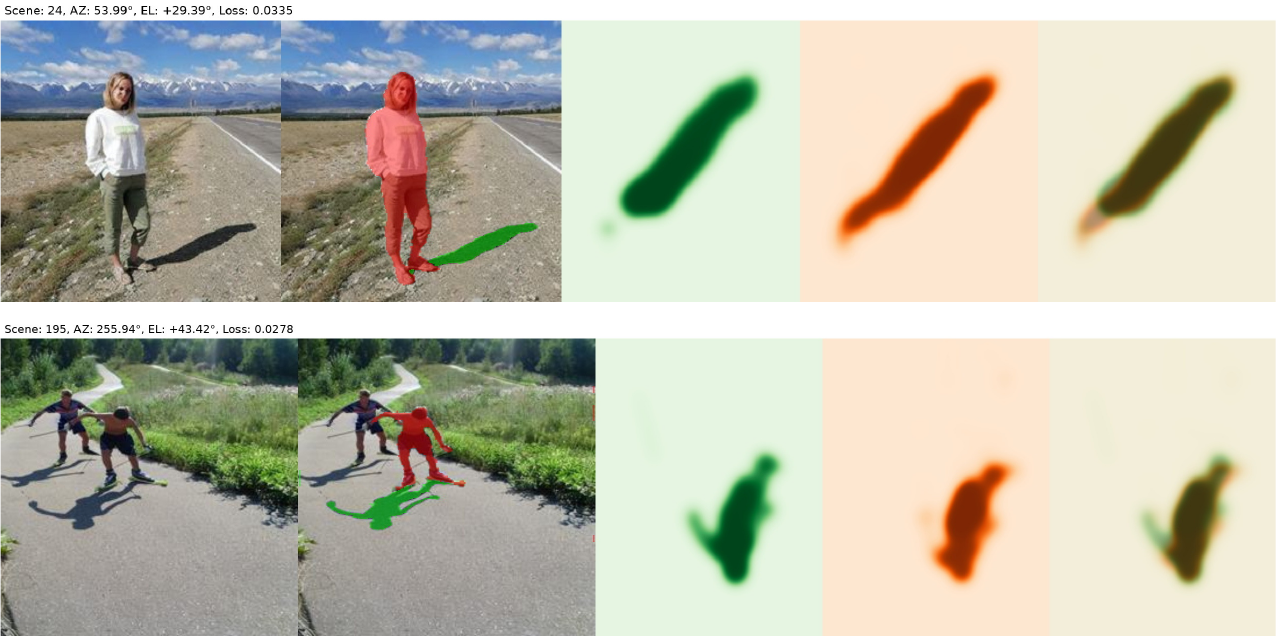}
   \caption{\textbf{Valid approximate light.} Examples where the automatically estimated light direction produces a rendered shadow density that closely matches the observed foreground shadow mask. The cast shadows from the object points align well with the observed shadow location and shape, indicating a plausible approximate light direction.}
   \label{fig:good_light}
\end{figure}

\begin{figure}[!t]
   \centering
   \includegraphics[width=\linewidth]{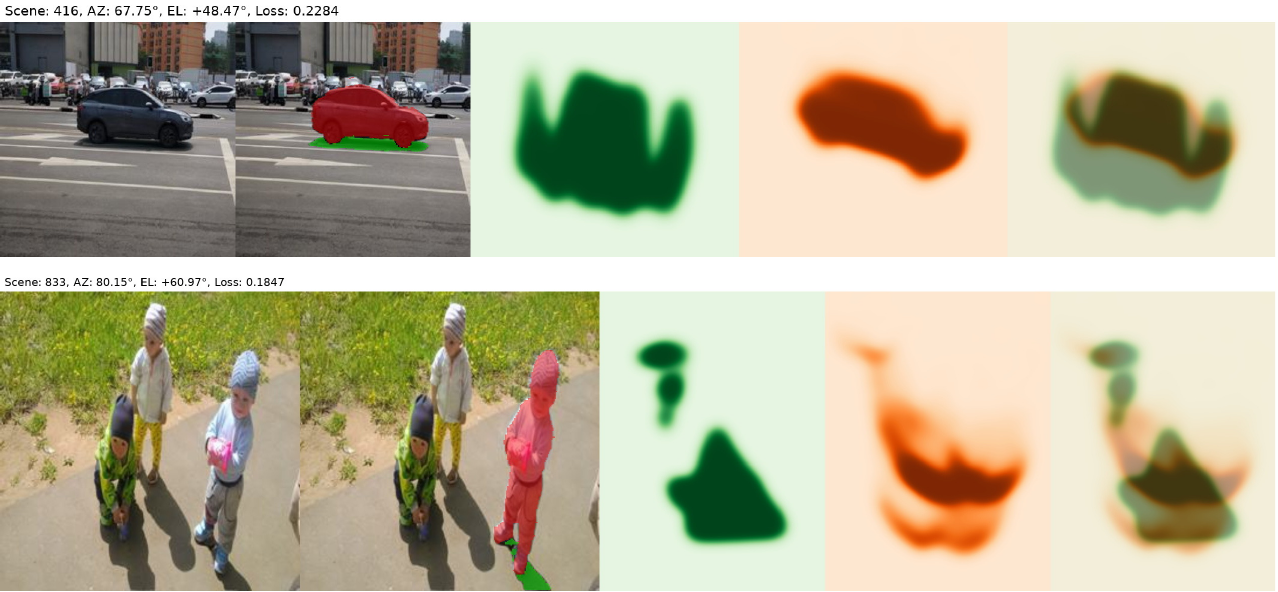}
   \caption{\textbf{Invalid approximate light.} Examples rejected by our manual check. Here, the approximate light either yields cast shadows that noticeably deviate from the observed shadow region (in location or shape), or the observed shadow is too partial to support a reliable elevation estimate. Such cases are discarded from the supervision.}
   \label{fig:bad_light}
\end{figure}

\paragraph{Shadow casting.}
Assuming a single directional light, we parameterize the light direction, pointing toward the source, by azimuth $\phi$ and elevation $\theta$:
\begin{equation}
\boldsymbol{\ell}(\phi,\theta)
=[-\cos\phi\cos\theta,-\sin\theta,-\sin\phi\cos\theta].
\end{equation}

For each occluder point $\mathbf{x}_i$, we cast a ray along $\mathbf{d}=-\boldsymbol{\ell}$ and intersect it with the receiver plane:
\begin{equation}
\mathbf{y}_i=\mathbf{x}_i+t_i\mathbf{d},\quad
t_i=\frac{(\bar{\mathbf{r}}-\mathbf{x}_i)\cdot\mathbf{n}}{\mathbf{d}\cdot\mathbf{n}} .
\end{equation}

Rays with $\lvert \mathbf{d}\cdot\mathbf{n}\rvert$ near zero or $t_i \le 0$ are treated as invalid. Valid intersections are projected to 2D receiver-grid coordinates, denoted by $\mathbf{q}_i$.

\paragraph{Shadow density rendering.}
The cast points are rendered onto a fixed grid using Gaussian splatting:
\begin{equation}
D_{\boldsymbol{\ell}}(\mathbf{p})=\sum_i w_i\,G_\sigma(\mathbf{p}-\mathbf{q}_i),
\end{equation}
producing a soft shadow density map. 
The density is then converted into a soft binary shadow estimate by applying a percentile-based threshold $\tau$ and logistic flattening:
\begin{equation}
R_{\boldsymbol{\ell}}=\sigma(\gamma(D_{\boldsymbol{\ell}}-\tau)),
\end{equation}
where $\gamma$ controls the transition sharpness. The observed shadow mask is processed in the same way to obtain a target map $R_S$.

\paragraph{Light optimization.}
We estimate the light direction by minimizing the discrepancy between rendered and observed shadows using a soft Dice loss together with an out-of-bounds penalty $\mathcal{L}_{oob}$ that discourages cast points from falling outside the receiver ROI:
\begin{equation}
\mathcal{L}=1-
\frac{2\langle R_{\boldsymbol{\ell}},R_S\rangle}
{\|R_{\boldsymbol{\ell}}\|_1+\|R_S\|_1+\varepsilon}
+\lambda_{oob}\mathcal{L}_{oob}.
\end{equation}
We first perform a coarse grid search over $(\phi,\theta)$ and then refine the best candidates with gradient-based optimization. The final estimate
\begin{equation}
\hat{\boldsymbol{\ell}}=\boldsymbol{\ell}(\hat{\phi},\hat{\theta})
\end{equation}
is used as pseudo supervision for training the light predictor.

\paragraph{Manual verification.}
After automatic estimation, we visualize the rendered shadow density and compare it with the observed shadow mask (Figure~\ref{fig:good_light}). Estimates that do not adequately explain the observed shadow location and shape are rejected, as are cases with observed shadows too partial for reliable estimation (Figure~\ref{fig:bad_light}). After filtering, we retain valid directions for 17,104 of the 21,575 scenes, covering 22,364 image tuples: 21,754 training tuples and 610 test tuples. Training-split labels are used only to supervise light-related losses, while test-split labels are used only for evaluation and appendix diagnostics. The pipeline runs offline; standard inference requires no pseudo-labels.

\paragraph{Evaluation without valid pseudo-labels.}
The remaining 140 test tuples have no valid light-direction pseudo-label. We use label availability only to define a diagnostic subset and evaluate our method and GPSDiffusion~\citep{zhao2025shadow} on these tuples under the same best-of-five protocol. As shown in Table~\ref{tab:no_light}, our method outperforms GPSDiffusion across all six metrics. This indicates that the gains are not confined to cases for which the offline pipeline can recover a valid light direction.

\begin{table}[!t]
\centering
\caption{Performance on the 140 DESOBAV2 test tuples without valid light-direction pseudo-labels. GR/LR, GS/LS, and GB/LB denote global/local RMSE, SSIM, and BER, respectively. All results use best-of-five selection. Best scores are in \textbf{bold}.}
\label{tab:no_light}

\small
\setlength{\tabcolsep}{4.0pt}
\renewcommand{\arraystretch}{1.08}

\begin{tabular}{@{}l c c c c c c@{}}
\toprule
\textbf{Method} &
\textbf{GR}$\downarrow$ &
\textbf{LR}$\downarrow$ &
\textbf{GS}$\uparrow$ &
\textbf{LS}$\uparrow$ &
\textbf{GB}$\downarrow$ &
\textbf{LB}$\downarrow$ \\
\midrule
GPSDiffusion~\citep{zhao2025shadow}
& 6.16 & 39.50 & 0.9692 & 0.5029 & 0.1513 & 0.2969 \\

Ours
& \textbf{4.28}
& \textbf{33.83}
& \textbf{0.9749}
& \textbf{0.5637}
& \textbf{0.1261}
& \textbf{0.2497} \\
\bottomrule
\end{tabular}
\end{table}

\section{More Qualitative Results}
\label{sec:morequal}

\begin{figure}[!t]
\centering
\includegraphics[width=\linewidth]{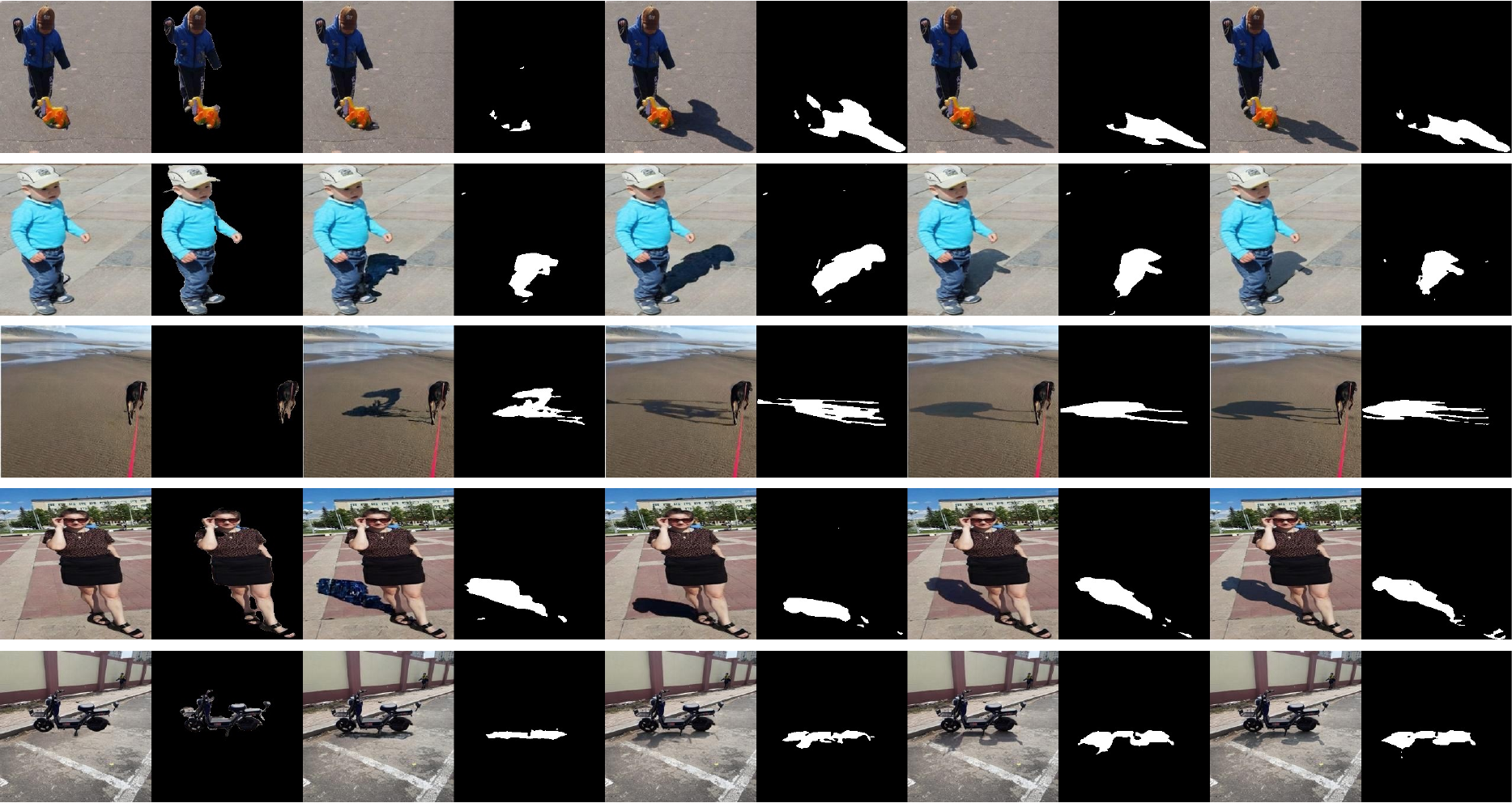}
\\ [1pt]
\includegraphics[width=\linewidth]{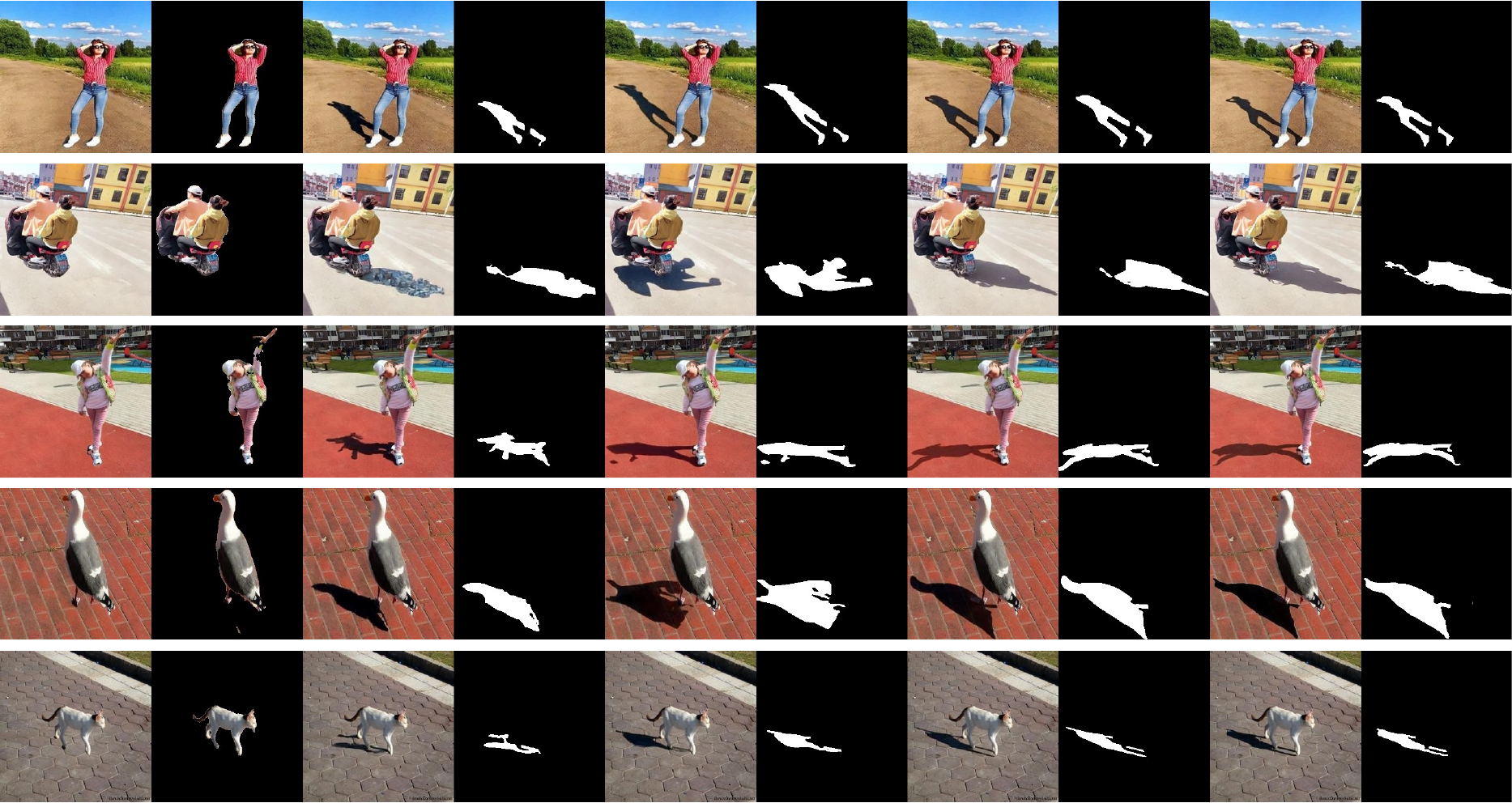}
{\small
  \makebox[0.19\linewidth]{Img. \& Obj.}%
  \makebox[0.19\linewidth]{SGDiffusion}%
  \makebox[0.21\linewidth]{GPSDiffusion}%
  \makebox[0.19\linewidth]{Ours}%
  \makebox[0.19\linewidth]{GT Img. + Mask}%
}
\caption{\textbf{Additional qualitative results in the BOS-free setting.}
We compare the generated images and predicted shadow masks from SGDiffusion~\citep{liu2024shadow}, GPSDiffusion~\citep{zhao2025shadow}, and our method with the ground truth. Across these examples, our method produces shadows with more accurate placement, shape, and extent.}
\label{fig:sup1}
\end{figure}

\begin{figure}[!t]
\centering
\includegraphics[width=\linewidth]{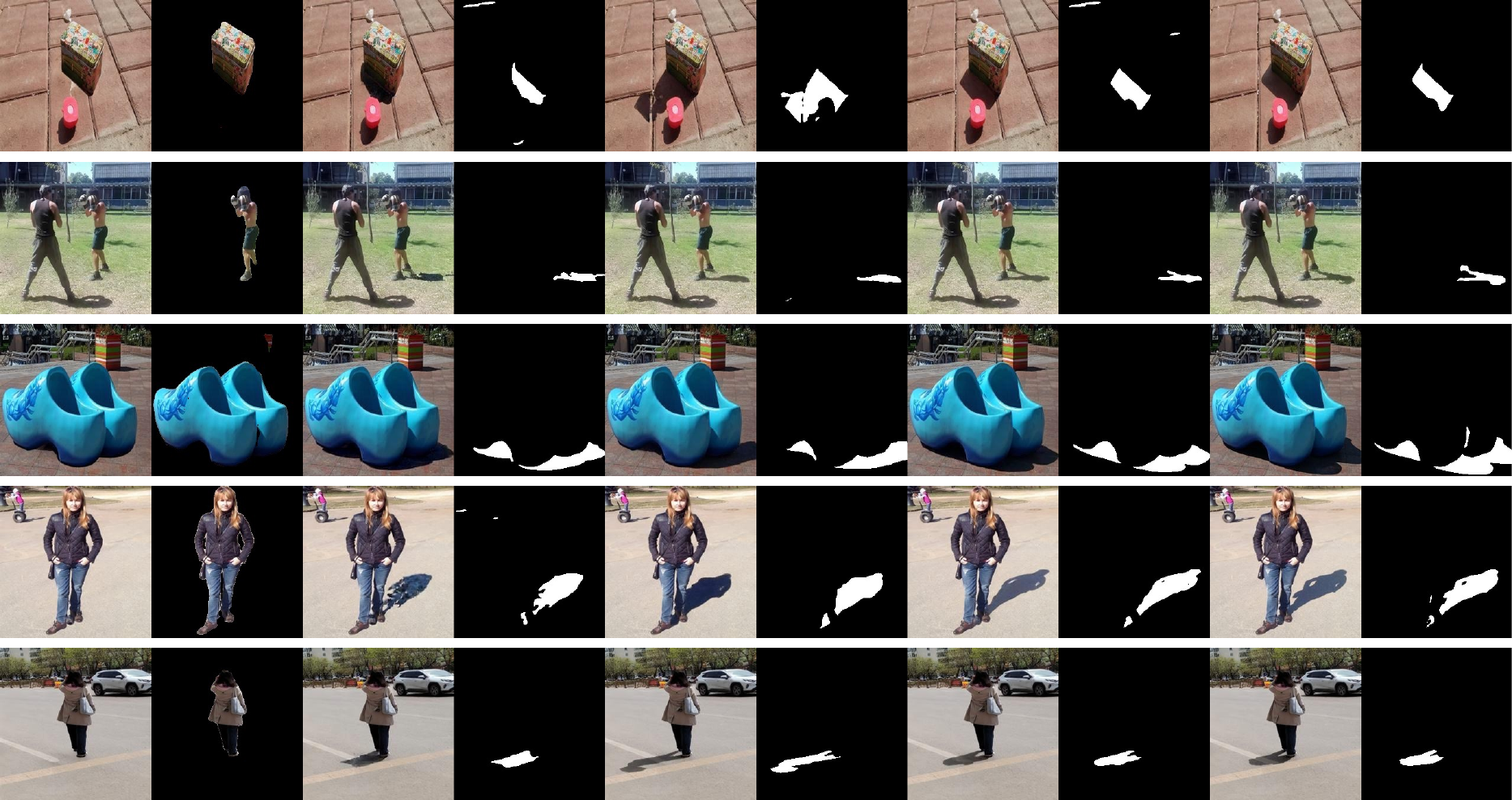}
\\ [1pt]
\includegraphics[width=\linewidth]{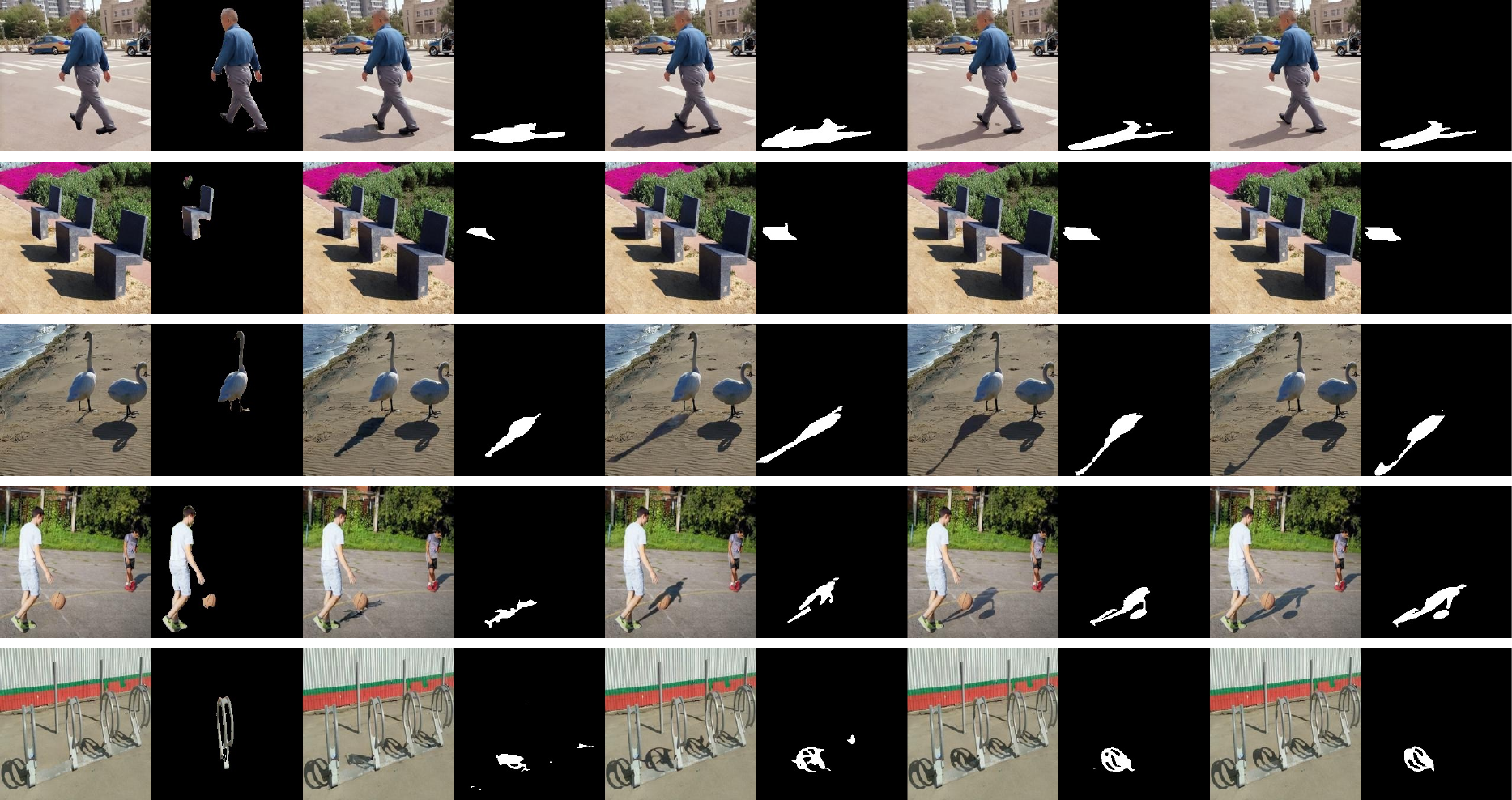}
{\small
  \makebox[0.19\linewidth]{Img. \& Obj.}%
  \makebox[0.19\linewidth]{SGDiffusion}%
  \makebox[0.21\linewidth]{GPSDiffusion}%
  \makebox[0.19\linewidth]{Ours}%
  \makebox[0.19\linewidth]{GT Img. + Mask}%
}
\caption{\textbf{Additional qualitative results in the BOS setting.}
We compare the generated images and predicted shadow masks from SGDiffusion~\citep{liu2024shadow}, GPSDiffusion~\citep{zhao2025shadow}, and our method with the ground truth. Across these examples, our method produces shadows with more accurate placement, shape, and extent.}
\label{fig:sup2}
\end{figure}

Figures~\ref{fig:sup1} and~\ref{fig:sup2} provide additional qualitative comparisons in the BOS-free and BOS settings, respectively. For each scene, we show the input image and foreground object mask, the generated images and predicted shadow masks from SGDiffusion~\citep{liu2024shadow}, GPSDiffusion~\citep{zhao2025shadow}, and our method, and the corresponding ground truth.

\end{document}